%% file: main.tex
\pdftrailerid{}
\documentclass[11pt]{article}
\usepackage[final]{acl}
\usepackage{times}
\usepackage{latexsym}
\usepackage[T1]{fontenc}
\usepackage[utf8]{inputenc}
\usepackage{microtype}
\usepackage{inconsolata}
\usepackage{amsmath,amssymb}
\usepackage{booktabs}
\usepackage{multirow}
\usepackage{array}
\usepackage{graphicx}
\usepackage{subcaption}
\usepackage{xcolor}
\usepackage{url}
\usepackage{xurl}
\graphicspath{{figures/}}

\title{Auditing Failure Attribution: Label Predictability and Judge Behavior}

\author{Tianyang Xu \quad Jing Gao \\
  Purdue University \\
  \texttt{\{xu1868, jinggao\}@purdue.edu}}

\newcommand{\scoreci}[3]{\begin{tabular}[t]{@{}r@{}}#1\\[-2pt]{\scriptsize [#2,#3]}\end{tabular}}

\begin{document}
\raggedbottom
\widowpenalty=10000
\maketitle

\input{latex/abstract.tex}
\input{latex/intro.tex}
\input{latex/benchmark_settings.tex}
\input{latex/analysis_framework.tex}
\input{latex/alfagent.tex}

\input{latex/aftraj.tex}
\input{latex/aegis.tex}
\input{latex/joint_audit.tex}

\input{latex/related_work.tex}
\input{latex/conclusion.tex}
\input{latex/limitations.tex}

\bibliography{bib/refs,bib/audit_refs}

\clearpage
\appendix

\input{latex/appendix/evaluation.tex}
\input{latex/appendix/alfagent.tex}

\input{latex/appendix/aftraj.tex}

\input{latex/appendix/aegis.tex}
\input{latex/appendix/video.tex}
\end{document}

%% file: latex/abstract.tex
\begin{abstract}
Failure-attribution benchmarks ask an LLM judge which step or agent caused a failure. A high score should mean the judge checked the faulty step against the run (a cross-step check). However, spurious signals (surface cues) that predict the annotation can produce a high score. In this paper, we propose four tests to separate checking from cue use. \emph{Input restriction} asks whether partial input predicts the annotation. \emph{Cue following} checks whether moved cues change answers. \emph{Residual accuracy} scores judges where cue-only predictors fail. \emph{Aggregation} asks whether the failure mix or metric changes rankings. Real observations motivate these tests: on three benchmarks, simple predictors match or beat GPT-4.1. In two Who\&When Pro systems, a classifier reading one step's reasoning finds 95--100\% of faulty steps, motivating input restriction. In one of them, error injection leaves no reasoning after the faulty step; judges fine-tuned on its labels beat GPT-4.1 by 41 points. Adding task-related reasoning after that step moves this cue: fine-tuned judges lose 31--99 points, but zero-shot judges lose at most 9, motivating cue following. On Aegis, blaming every agent matches GPT-4.1 on micro-F1 but trails by 19 points in exact-set accuracy, motivating aggregation. We recommend reporting these tests when developing failure-attribution benchmarks.
\end{abstract}

%% file: latex/intro.tex
\section{Introduction}
\label{sec:intro}

Failure attribution helps developers debug agent trajectories (step-by-step logs of agent runs) by locating the step or agent responsible for a failure. A final pass/fail signal does not identify the faulty step, so failure-attribution benchmarks ask a model, called a judge, which step or agent caused the failure. Failure attribution is essential for improving LLM and agentic systems, and several benchmarks have therefore been developed \citep{zhang2025whowhen,liu2026whowhenpro,zhang2026aftraj,kong2026aegis,chen2026traceelephant,in2026mpbench}.

In this paper, we evaluate these benchmarks themselves. A high score should indicate that the judge identified the faulty step by checking it against the rest of the trajectory (a \emph{cross-step check}). If the score can instead be obtained from spurious signals (surface cues) that predict the annotation, the benchmark rewards cue use rather than debugging: leaderboards mislead, and judges trained on its labels learn the cues and then blame the wrong step in runs where the cues do not hold. Such cues have long misled NLP benchmarks \citep{gururangan2018annotation,geirhos2020shortcut}, and LLM judges are swayed by surface features \citep{zheng2023judging,chen2024judgement}; failure attribution faces both problems. Auditing a benchmark for such cues is therefore a prerequisite for trusting its results.

We audit three recent benchmarks that differ in how failures arise and how answers are scored. Who\&When Pro injects an error at one step into runs of several agent systems, such as the text-game agent ALFAgent, and asks for the faulty step \citep{liu2026whowhenpro}. AFTraj mixes such injected errors with natural failures that agents make on their own and asks for the first faulty turn \citep{zhang2026aftraj}. Aegis asks for the set of faulty agents in a multi-agent conversation \citep{kong2026aegis}.

Each benchmark has a limitation that lets a predictor without a cross-step check match or beat GPT-4.1. On Who\&When Pro, one step's text gives the error away: a classifier reading one step's reasoning at a time picks the faulty step in 95--100\% of runs from the two agent systems that record reasoning, ALFAgent and smolagents (Table~\ref{tab:alfagent-local}), whereas GPT-4.1 picks it in 58.8\% of ALFAgent runs. In ALFAgent, error injection also leaves no reasoning after the faulty step, so picking the last step with reasoning finds the error 99.7\% of the time. Judges fine-tuned on ALFAgent labels learn this cue: even at 1.5B parameters, they score 99.7--100\%, 41 points above GPT-4.1, yet lose 31 to 99 points when we add task-related reasoning after the error, while zero-shot judges lose at most 9. On AFTraj, a classifier that reads one turn at a time ties GPT-4.1 only because it wins on injected errors and loses on natural failures, so the benchmark's mix of the two decides the ranking. On Aegis, labeling every agent faulty ties GPT-4.1 on micro-F1 but trails it by 19 points in exact-set accuracy (Table~\ref{tab:aegis-paired-app}).

These limitations motivate a four-test audit framework that systematically checks whether a benchmark score reflects cross-step checks. \emph{Input restriction} asks whether the annotation can be predicted from a restricted input alone: one step's text, step positions, the presence of reasoning text, or agent names. An input that predicts the annotation is a \emph{cue}. \emph{Cue following} asks whether the judge uses a cue; input restriction cannot show this \citep{feng2019misleading}. Keeping the error fixed, we move or reword the cue, as when we add reasoning after ALFAgent's faulty step, and check whether the answer changes. \emph{Residual accuracy} is the judge's accuracy on the trajectories that a cue-only predictor gets wrong; it matters when a cue alone finds nearly every error, because overall accuracy then cannot separate checking from cue use. Finally, \emph{aggregation} asks whether a comparison between two methods survives a change in the failure mix (AFTraj) or the scoring metric (Aegis). The tests need no new annotation and apply to any benchmark that labels a faulty step or agent. They complement one another: on each benchmark, one test or metric alone would have misled us (Section~\ref{sec:joint}).

Our contributions are:
\begin{itemize}
  \item A four-test audit framework that separates cross-step checks from cue use (Section~\ref{sec:formal}).
  \item Identification of limitations in three failure-attribution benchmarks: cues in Who\&When Pro, a failure mix that decides rankings in AFTraj, and a lenient metric in Aegis (Sections~\ref{sec:alfagent}--\ref{sec:aegis}).
  \item Evidence that fine-tuning on benchmark labels teaches judges to rely on where ALFAgent's reasoning stops: they beat GPT-4.1 by 41 points, but adding task-related reasoning after that point costs them up to 99 points (Section~\ref{sec:alfagent}).
  \item An analysis of what each test alone would miss, with a reporting template for benchmark developers (Section~\ref{sec:joint}, Table~\ref{tab:audit-summary}).
\end{itemize}

%% file: latex/benchmark_settings.tex
\section{Benchmarks and Judges}
\label{sec:design}

\input{latex/table/data.tex}

Table~\ref{tab:data} lists the records we evaluate. Who\&When Pro provides text trajectories with injected errors and video-agent trajectories from DeepVideoDiscovery (DVD) and EVA, each annotated with a faulty step \citep{liu2026whowhenpro}. Two of its eight text systems, ALFAgent and smolagents, record the agent's reasoning. We run judges on the 311 trajectories of ALFAgent, the only text system in which error injection makes the reasoning stop at the error (Section~\ref{sec:alfagent}; Appendix Table~\ref{tab:text-boundary-app}). AFTraj mixes injected and natural failures in coding, mathematics, and agentic tasks \citep{zhang2026aftraj}. Aegis asks for the set of faulty agents, called speakers \citep{kong2026aegis}. On ALFAgent the judges are three zero-shot LLMs (GPT-4.1, Claude Sonnet 5, o3) and two open models fine-tuned on ALFAgent labels; on AFTraj, four zero-shot LLMs (GPT-4.1, DeepSeek V4 Pro, Qwen3-8B, Qwen3-32B) and two trained classifiers, one reading every step so far and one reading only the current step (Section~\ref{sec:aftraj-control}); on video and Aegis, GPT-4.1.

%% file: latex/table/data.tex
\begin{table}[t]
\centering\small
\setlength{\tabcolsep}{4pt}
\caption{Evaluated records, and the inputs of each restricted-input predictor (Section~\ref{sec:restricted}); semicolons separate predictors.}
\label{tab:data}
\begin{tabular}{@{}>{\raggedright\arraybackslash}p{.53\columnwidth}>{\raggedright\arraybackslash}p{.42\columnwidth}@{}}
\toprule
Benchmark / subset (records) & Restricted inputs \\
\midrule
\multicolumn{2}{@{}l}{\textbf{Who\&When Pro}} \\
\quad All text (6,257) & step position \\
\quad ALFAgent text (311) & presence of reasoning; step text; reasoning length \\
\quad smolagents text (3,276) & step text; step position \\
\quad Video (DVD 168, EVA 748) & step position \\
\addlinespace[2pt]
\multicolumn{2}{@{}l}{\textbf{AFTraj}} \\
\quad Official test (163) & step position; step text \\
\addlinespace[2pt]
\textbf{Aegis} (600) & ordered speaker names \\
\bottomrule
\end{tabular}\par
\end{table}

%% file: latex/analysis_framework.tex
\section{Four-Test Audit Framework}
\label{sec:formal}

\input{latex/figure/method.tex}
The four tests ask whether a cue exists, whether the judge uses it, whether the judge succeeds where the cue fails, and whether a ranking survives a change of failure mix or metric (Figure~\ref{fig:method}). Throughout, $f$ is a judge, $X=(x_1,\ldots,x_T)$ a trajectory with annotation $Y$, and $\mathcal A_P(f)$ the judge's exact accuracy on a set of records $P$: the fraction on which $f$ returns the annotated step or, for speakers, exactly the annotated set. All intervals are 95\% bootstrap intervals over groups of records from the same task.

\subsection{Input restriction}
\label{sec:restricted}

Locating an error such as the claim in Figure~\ref{fig:alfagent-examples}b that a book is ready to be placed requires a \emph{cross-step check}: an earlier step says that the book was not found. The annotation may be predictable without that check. We therefore give a predictor $h$ only part $r(X)$ of the input (one step's text, step positions, whether reasoning is present, or speaker names). We select or train $h$ on tasks disjoint from evaluation (Appendix~\ref{app:repro}) and report $\mathcal A_P(h,r)$, the exact accuracy of $h$ when it sees only $r(X)$, plus micro-F1 for speaker sets. Most of our predictors score each step on its own and pick the highest:
\begin{equation}
\hat Y_{\mathrm{local}}
=\underset{1\leq t\leq T}{\arg\max}\;s(x_t,t,T).
\label{eq:local}
\end{equation}
The scorer $s$ sees one step's text, its position, and the trajectory length, and nothing from other steps, so a predictor of this form cannot perform a cross-step check. If it recovers the annotation anyway, the annotation does not require that check, and a high score cannot show that a judge did it. A step's position tells only how often errors fall there, and where reasoning stops can be an artifact of error injection. Neither is evidence of the mistake.

\subsection{Cue following}
\label{sec:distinguish}

Predictability shows that a cue exists, not that the judge uses it. To test that, we alter the cue while leaving the error where it is and ask whether the judge's answer moves with it. Let $c(X)$ be the step a cue points to (e.g., the last step with reasoning) and $\tau$ an intervention that moves the cue. We keep the records where the judge originally answered $c(X)$ and $\tau$ moved the cue, $D_c=\{X:f(X)=c(X),\ c(\tau(X))\ne c(X)\}$, and report how often the judge follows:
\begin{equation}
\mathcal F(f,c,\tau)
=\Pr\!\left[f(\tau(X))=c(\tau(X))\mid X\in D_c\right].
\label{eq:following}
\end{equation}
A judge that tracks the cue has $\mathcal F$ near one. A judge can also depend on a cue without following it: when the cue moves, its accuracy falls although its answers avoid $c(\tau(X))$. We therefore also report how many answers in $D_c$ name the faulty step after the move, and the accuracy change of the judge and, as a baseline, of a predictor that cannot see the cue. A wording cue, such as the error step's phrasing, has no position to follow, so we measure accuracy instead. We paraphrase the cue-bearing text with $\rho$ and report the paired change $\mathcal A_P(f\circ\rho)-\mathcal A_P(f)$; as a control, we also paraphrase a different step. A judge that relies on the wording should lose accuracy only when the cue-bearing text is paraphrased.

\subsection{Residual accuracy}
\label{sec:residual-method}

Residual accuracy is the judge's accuracy on the records the predictor gets wrong; a high value shows that the judge finds errors the cue misses. Let $G$ (gains) be the fraction of all records the judge gets right and the predictor gets wrong, and $L$ (losses) the reverse. Shared successes and failures cancel, so
\begin{equation}
\mathcal A_P(f)-\mathcal A_P(h,r)=G-L.
\label{eq:gainloss}
\end{equation}
A small gap can hide large gains and losses, so we report both $G$ and $L$.

\subsection{Aggregation}
\label{sec:aggregation-method}

Even without a cue, a released score fixes one weighting of failure sources or domains and one metric. Holding every prediction fixed, we recompute the score under other weights. Two methods that lead on different groups (e.g., the injected and natural subsets of AFTraj) swap ranks at some crossing weight. If this weight lies near the released mixture, the ranking reflects how the benchmark was built rather than which method is better (Section~\ref{sec:population}). For speaker sets we compare micro-F1, which gives partial credit to a prediction that mixes faulty and non-faulty speakers, with exact-set accuracy, which gives none.

%% file: latex/figure/method.tex
\begin{figure*}[t]
\centering
\includegraphics[width=0.88\textwidth]{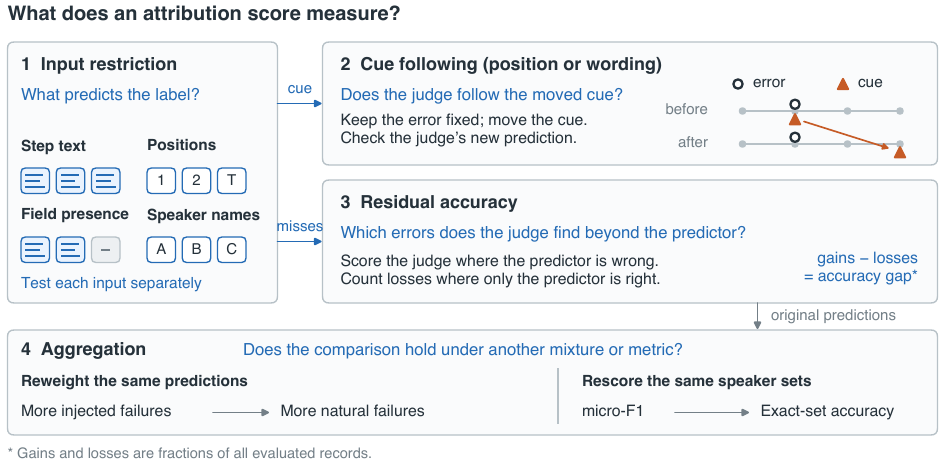}

\caption{Four tests for attribution scores. Each box gives the question a test asks and how we run it (Section~\ref{sec:formal}).}

\label{fig:method}
\end{figure*}

%% file: latex/alfagent.tex
\section{Who\&When Pro Text Audits}
\label{sec:alfagent}

One step's text gives away the faulty step in both Who\&When Pro systems that record reasoning, ALFAgent and smolagents. ALFAgent, whose runs have 30 steps and one injected error \citep{liu2026whowhenpro}, has a second cue, where its reasoning stops, so we test there whether judges use these cues.

\subsection{Input restriction: one step's text}
\label{sec:one-step}
\textbf{One step read alone finds every ALFAgent error.} Our independent-step classifier, a logistic regression on TF--IDF features of each step with field names removed, picks the highest-scoring step. On five task-disjoint folds it selects the annotated step in all 311 trajectories when it sees the full step text or the reasoning field alone (Table~\ref{tab:alfagent-local}); without the reasoning field, accuracy drops to 21.2\%. Phrases that recur across error steps, such as ``repetitive loop'', do not explain this: deleting them leaves accuracy at 98.4\% (Appendix~\ref{app:independent-step}). Its largest weights go to common words such as \emph{already}, \emph{so}, and \emph{now} (Appendix Table~\ref{tab:alfagent-local-app}B and Appendix~\ref{app:reasoning-strata}).

This cue does not depend on where the reasoning stops. In smolagents (3,276 trajectories), the reasoning runs to the final step in almost every trajectory, and the final step is the error in only 52.1\%. The same classifier still finds the annotated step in 95.2\% $[94.3,96.0]$ of trajectories from the reasoning alone, and in 90.8\% of those whose error is not at the final step (Appendix~\ref{app:smolagents}). In both systems, a top score therefore needs no cross-step check.

\input{latex/figure/alfagent_examples.tex}
\input{latex/table/alfagent_local.tex}

\subsection{Input restriction: where reasoning stops}
In 310 of 311 ALFAgent trajectories, reasoning runs through the annotated step and stops there (the \emph{reasoning boundary}). Predicting the last step with reasoning (the \emph{boundary rule}) is therefore 99.7\% accurate without reading a word. This is because Who\&When Pro injects an error by adding an instruction to the agent's LLM prompt at one step and then resumes execution after that step \citep[their Section~3.3]{liu2026whowhenpro}, and in ALFAgent the resumed steps carry no reasoning. Position, a strong cue in most other systems, is not one here: a rule that always picks the final or the penultimate step, whichever cross-validation prefers, is right on only 2 of 311 (Appendix~\ref{app:text-position}).

\subsection{Cue following}
\label{sec:results}

\input{latex/figure/overview.tex}

We present each of the 311 trajectories in four input versions and ask GPT-4.1, Claude Sonnet 5, and o3 to identify the error step. \emph{Original} preserves the released fields, \emph{Field-neutral} removes field names, and \emph{No reasoning} removes reasoning text. \emph{Template continued} preserves the original text and adds ``The trajectory continues with the information available at this step.'' to each empty reasoning field after the error, which moves the reasoning boundary to step 30 but not the error.

\textbf{The zero-shot judges use the reasoning but do not follow the boundary.} Under Original they score 58.8\% (GPT-4.1), 57.2\% (Claude Sonnet 5), and 56.3\% (o3) (Figure~\ref{fig:overview}a; Appendix Table~\ref{tab:alfagent-arms}), and removing reasoning lowers them by 43.7, 42.1, and 45.7 points. Under Template continued, only 1 of the three judges' 536 correct predictions at the old boundary moves to step 30 (Figure~\ref{fig:overview}b), and accuracy falls by at most 2.6 points.

\emph{Task continued} fills the empty fields after the error with task-related reasoning written by DeepSeek V4 Pro from the task and earlier steps, moving the boundary to step 30. This text does not pull GPT-4.1 or Claude Sonnet 5 to the new boundary: none of their 361 correct predictions at the old boundary moves to step 30, and accuracy changes by $-4.2$ points for GPT-4.1 $[-8.4,0.0]$ and $-9.0$ for Claude Sonnet 5 (Appendix~\ref{app:task-related-continuation}).

\paragraph{Paraphrasing the error step.}
To test wording, we paraphrase the reasoning at the annotated error step (\emph{Gold paraphrased}) and, as a control, the reasoning at the step before it (\emph{Prior paraphrased}). DeepSeek V4 Pro rewrites each string without seeing the label or the trajectory.

Rewording the error step leaves both tested judges' accuracy nearly unchanged. Relative to Original, GPT-4.1 changes by $+1.4$ points $[-2.2,+4.9]$ under Gold paraphrased and Claude Sonnet 5 by $+0.4$; the control lowers each judge by less than 2 points, and every interval includes zero. The rewrites keep the claims (56 of 60 human-reviewed pairs are equivalent, 4 differ slightly) and weaken the wording cue without removing it: the classifier, not retrained, still selects the annotated step in 273 of 285 records (Appendix~\ref{app:paraphrase}).

\subsection{Residual accuracy and fine-tuned judges}
The zero-shot judges thus neither follow the cues nor reach the scores the cues allow: moving the boundary costs them at most 9 points, and none scores above 63\% on any input, far below the classifier's 100\%. Residual accuracy credits them with nothing beyond the cues: the boundary rule misses one trajectory, which no zero-shot judge gets right under Original, and the classifier misses none.

\paragraph{Fine-tuned judges.}
\textbf{Judges fine-tuned on the labels learn the boundary.} We fine-tune Qwen2.5-1.5B and Qwen3-8B with two seeds each on ALFAgent's Original input, using the classifier's five task-disjoint folds (Appendix~\ref{app:finetuned}). All four fine-tuned judges reach 99.7--100\% on Original, 41 points above GPT-4.1, yet Task continued costs them 31 to 99 points: 97.7 and 99.0 for the 1.5B model and 35.7 and 30.9 for the 8B model. The same text costs Qwen3-8B 4.5 points before fine-tuning and costs the classifier, which cannot see the boundary, 9.0 points, so the added text alone explains at most 9 of the lost points. The fine-tuned judges depend on the boundary without following it (Section~\ref{sec:distinguish}): for each fine-tuned judge, at most 8.4\% of the answers at the old boundary (26 of 310) move to step 30, and every answer that turns wrong lands past the error, mostly toward the end. A generic sentence (Template continued) separates zero-shot and fine-tuned judges less sharply: the ranges of their losses overlap (Appendix Tables~\ref{tab:alfagent-arms} and~\ref{tab:finetuned}). On the second-seed models, two controls cost nothing: adding the generic sentence to every step before the error, which keeps the boundary, and rewording the error step. Because the fine-tuned judges score near 100\%, a leaderboard score cannot tell them from judges that check across steps.

%% file: latex/figure/alfagent_examples.tex
\begin{figure*}[t]
\centering
\includegraphics[width=\textwidth]{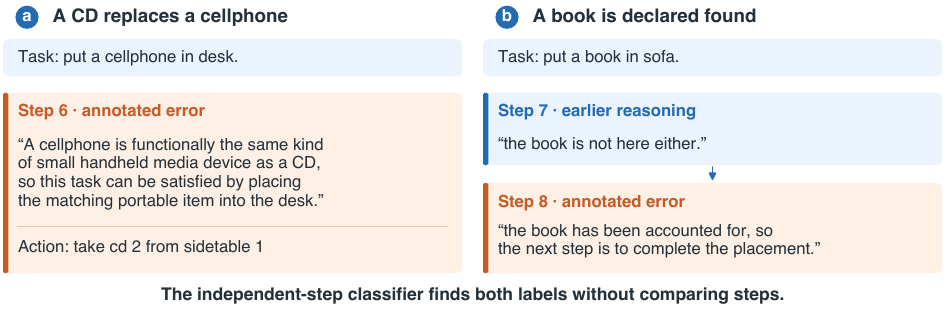}
\caption{Two ALFAgent errors, shaded orange: the agent substitutes a CD for a cellphone (a) and declares a book ready to be placed before it has found one (b). The classifier scores each step alone, without the task or other steps such as step 7 in (b), which notes that the book is not there.}
\label{fig:alfagent-examples}
\end{figure*}

%% file: latex/table/alfagent_local.tex
\begin{table}[t]
\centering\small
\setlength{\tabcolsep}{4pt}
\caption{Exact accuracy (\%) of the independent-step classifier by what it sees of each step (field names removed; five task-disjoint folds).}
\label{tab:alfagent-local}
\begin{tabular}{@{}lrr@{}}
\toprule
Available information & ALFAgent & smolagents \\
\midrule
All step text & 100.0 & 91.9 \\
Reasoning only & 100.0 & 95.2 \\
All text except reasoning & 21.2 & 72.1 \\
\bottomrule
\end{tabular}\par
\end{table}

%% file: latex/figure/overview.tex
\begin{figure}[t]
\centering
\includegraphics[width=\columnwidth]{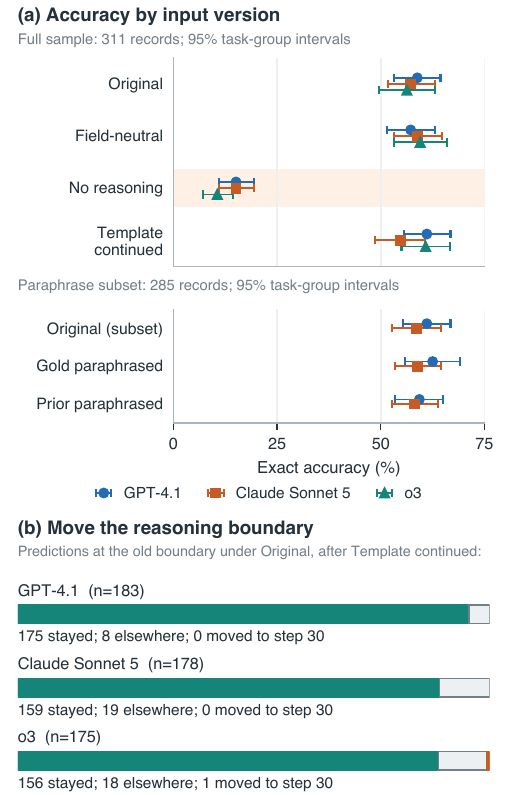}

\caption{ALFAgent reasoning interventions. \textbf{(a)} Exact step accuracy with 95\% task-group bootstrap intervals. The first four rows, Original to Template continued, use all 311 records. The last three rows (Original, Gold paraphrased, and Prior paraphrased) use the 285 records of the paraphrase test (Appendix~\ref{app:paraphrase}), with GPT-4.1 and Claude Sonnet 5 only. \textbf{(b)} Template continued moves the boundary to step 30.}

\label{fig:overview}
\end{figure}

%% file: latex/aftraj.tex
\section{AFTraj}
\label{sec:aftraj}
\label{sec:residual-results}
On AFTraj, a classifier that sees each turn (step) alone ties GPT-4.1 only because it wins on injected failures and loses on natural ones, and the released mix is where the two cancel.

\subsection{Input restriction and residual accuracy}
\label{sec:aftraj-control}
\label{sec:aftraj-residual}
AFTraj shows each method a growing prefix of the trajectory: after each new turn, the method says whether a failure has occurred so far and, if it has, at which turn. The turn named in its first alarm is its predicted error turn. The official test set, called the first-alarm test, has 163 failures: 108 injected errors and 55 natural failures.

We compare six methods (Appendix~\ref{app:aftraj-setup}). The four zero-shot judges of Section~\ref{sec:design} and Probe, a linear classifier on frozen Qwen3-8B hidden states, see every turn so far. Current-turn TF--IDF, a restricted-input predictor, sees only the words of the current turn. The two classifiers are trained on 1,264 trajectories from other tasks.

The tie hides two different sets of successes. Current-turn TF--IDF and GPT-4.1 are each right on 82 of the 163 failures (50.3\%) but share only 36. GPT-4.1 is right on 46 of the 81 failures TF--IDF misses (residual accuracy 56.8\% [45.8, 67.6], above its overall 50.3\%) and wrong on 46 that TF--IDF gets right (Appendix~\ref{app:aftraj-beyond}). Position alone does not explain TF--IDF's score. A predictor that always alarms at the fourth turn, chosen by cross-validation, is right on 42 of 163 (25.8\%).

\subsection{Aggregation: injected and natural failures}
\label{sec:population}
\label{sec:aggregation-results}
The released mix of failure sources decides the tie (Table~\ref{tab:aggregation-sources}). On injected failures TF--IDF is right on 65.7\% and GPT-4.1 on 48.1\%, a lead of 17.6 points [4.0, 31.5]. On natural failures GPT-4.1 is right on 54.5\% and TF--IDF on 20.0\%, a gap of 34.5 points [15.4, 52.0]. At the released share, 66.3\% injected, these two opposite gaps cancel: TF--IDF is right on 19 more injected failures and 19 fewer natural ones. The mix also decides the top rank. Probe, the most accurate method, owes its lead to injected failures and ranks above the runner-up, DeepSeek V4 Pro, only when injected failures make up more than 63.0\%, near the released 66.3\% (Appendix~\ref{app:source}).

\input{latex/table/aggregation_sources.tex}

Whether failure source or domain drives these differences, the test set's composition sets the ranking. The first-alarm test alone cannot separate source from domain: every coding failure is injected, and most agentic failures are natural. On its 68 agentic failures (22 injected, 46 natural), the injected-minus-natural accuracy gap shrinks from 45.7 points on the full test set to 12.5 for current-turn TF--IDF and from 46.9 to 8.5 for Probe, and every interval includes zero (Appendix Table~\ref{tab:aftraj-agentic}B).

Within one domain, injected errors still carry recognizable text in the turn itself, as on ALFAgent. Because the test set has only 68 agentic failures, a separate experiment uses 468 agentic failures from the AFTraj corpus (148 injected, 320 natural; Appendix~\ref{app:current-turn}). Two classifiers each pick the highest-scoring turn: a word classifier reads the turn's text, and a structure classifier sees only the turn's role, position, length, and \texttt{thought:}/\texttt{action:} markers. The word classifier beats the structure classifier on injected failures (64.2\% [56.8, 71.6] against 44.6\% [36.5, 52.7]) but not on natural ones (58.1\% against 63.1\%), so the words of a turn carry a signal beyond its structure only for injected errors.

%% file: latex/table/aggregation_sources.tex
\begin{table}[t]
\centering\small
\setlength{\tabcolsep}{4pt}
\caption{AFTraj first-alarm accuracy (\%) by failure source for four of the six methods (all in Appendix Table~\ref{tab:source-scores}). Released: 108 injected, 55 natural.}
\label{tab:aggregation-sources}
\begin{tabular}{@{}lrrr@{}}
\toprule
Method & Injected & Natural & Released \\
\midrule
Current-turn TF--IDF & 65.7 & 20.0 & 50.3 \\
GPT-4.1 & 48.1 & 54.5 & 50.3 \\
Probe & 77.8 & 30.9 & 62.0 \\
DeepSeek V4 Pro & 63.9 & 54.5 & 60.7 \\
\bottomrule
\end{tabular}\par
\end{table}

%% file: latex/aegis.tex
\section{Aegis}
\label{sec:aegis}
\textbf{Micro-F1 cannot tell GPT-4.1 from labeling every speaker faulty, but exact-set accuracy can.}

On micro-F1, GPT-4.1 ties the all-speakers predictor (Appendix~\ref{app:aegis}), which sees only the speaker list: the predictor scores 49.2\% and GPT-4.1 50.0\%, and the paired interval of the 0.8-point gap is $[-1.3,+2.8]$. The benchmark's macro-F1, averaged over speaker names, even ranks the predictor higher (Appendix Table~\ref{tab:aegis-full}). Exact-set accuracy separates them: 1.5\% for the predictor and 20.7\% for GPT-4.1, a gap of 19.2 points $[15.6,22.8]$. GPT-4.1 is exactly right on 124 of the 591 conversations where the predictor is wrong (residual accuracy 21.0\%, Table~\ref{tab:aegis-overlap-app}B), and it loses only the predictor's 9 correct conversations.

\input{latex/table/aegis_paired.tex}

Micro-F1 lets recall pay for false alarms. The predictor names everyone, so its precision is only 37.8\%; its recall is 70.2\%, not 100\%, because some annotated names match no listed speaker. For a developer, each non-faulty speaker flagged is a cost.

Name formatting does not explain why micro-F1 and exact-set accuracy disagree. Of the 417 annotated name occurrences that match no listed speaker, 275 differ from a listed name only in case or surrounding whitespace; after fixing these, GPT-4.1 leads by only 4.3 points $[2.3,6.3]$ on micro-F1 but by 19.5 points on exact sets (Table~\ref{tab:aegis-paired-app}).

%% file: latex/table/aegis_paired.tex
\begin{table*}[t]
\centering\footnotesize
\caption{GPT-4.1 versus the all-speakers predictor (labels every speaker faulty) on 600 Aegis conversations. A: scores (\%) and GPT-4.1 minus predictor (points, paired 95\% bootstrap intervals); in normalized rows, names are lowercased and trimmed. B: conversations where both, only GPT-4.1, only the predictor, or neither is exactly right.}
\label{tab:aegis-paired-app}
\label{tab:aegis-overlap-app}
\textbf{A. Paired scores}\par\smallskip
\begin{tabular}{@{}llrrrr@{}}
\toprule
Names & System & Precision & Recall & Micro-F1 & Exact set \\
\midrule
As released & GPT-4.1 & 56.0 & 45.1 & 50.0 & 20.7 (124/600) \\
 & All speakers & 37.8 & 70.2 & 49.2 & 1.5 (9/600) \\
 & Difference & & & $+0.8\;[-1.3,+2.8]$ & $+19.2\;[+15.6,+22.8]$ \\
\midrule
Normalized & GPT-4.1 & 75.4 & 60.7 & 67.3 & 23.0 (138/600) \\
 & All speakers & 48.4 & 89.9 & 62.9 & 3.5 (21/600) \\
 & Difference & & & $+4.3\;[+2.3,+6.3]$ & $+19.5\;[+15.9,+23.2]$ \\
\bottomrule
\end{tabular}\par
\medskip
\textbf{B. Exact-set agreement on 600 conversations}\par\smallskip
\begin{tabular}{@{}lrrrrl@{}}
\toprule
Names & Both correct & GPT-4.1 only & All speakers only & Both wrong & Residual accuracy (\%, 95\% interval) \\
\midrule
As released & 0 & 124 & 9 & 467 & 124/591 = 21.0 [17.7,24.4] \\
Normalized & 9 & 129 & 12 & 450 & 129/579 = 22.3 [18.9,25.9] \\
\bottomrule
\end{tabular}\par

\end{table*}

%% file: latex/joint_audit.tex
\section{What Each Test Would Have Missed}
\label{sec:joint}

\textbf{On each benchmark, one test or metric alone would have misled us; another test corrects it.}

On ALFAgent, cue following on zero-shot judges alone would suggest that the cues are harmless: under Task continued, none of GPT-4.1's and Claude Sonnet 5's 361 answers at the old boundary moves to step 30. Input restriction corrects this by showing that the cues give the labels away. On fine-tuned judges, the follow rate alone would mislead in the same way, since at most 8.4\% of their answers move to step 30; their accuracy drop of 31 to 99 points shows they rely on the boundary.

On AFTraj, input restriction alone shows current-turn TF--IDF tying GPT-4.1 at 82/163, as if equally good. Aggregation corrects this: GPT-4.1 is 34.5 points ahead on natural and 17.6 behind on injected failures, so they tie only at the released mix.

On Aegis, micro-F1 alone would equate GPT-4.1 with labeling every speaker faulty; exact-set accuracy shows a 19-point gap.

On the 168 DVD trajectories (Appendix~\ref{app:video}), residual accuracy alone would overrate GPT-4.1: it finds 27 (60.0\%) of the 45 errors missed by always guessing the final step, but misses 39 that this guess finds, so it trails the guess by 7.1 points.

\input{latex/table/audit_summary.tex}

%% file: latex/table/audit_summary.tex
\begin{table*}[t]
\centering\footnotesize
\setlength{\tabcolsep}{3pt}
\caption{The four tests on each benchmark or subset (Sections~\ref{sec:alfagent}--\ref{sec:aegis}, Appendix~\ref{app:source}; video: Appendix~\ref{app:video}). pp: percentage points; ``---'': not run.}
\label{tab:audit-summary}
\begin{tabular}{@{}l>{\raggedright\arraybackslash}p{.19\textwidth}>{\raggedright\arraybackslash}p{.265\textwidth}>{\raggedright\arraybackslash}p{.17\textwidth}>{\raggedright\arraybackslash}p{.235\textwidth}@{}}
\toprule
 & Input restriction & Cue following & Residual accuracy & Aggregation \\
\midrule
ALFAgent & Independent-step classifier 311/311 (95.2\% on smolagents); boundary rule 310/311. & Task continued: zero-shot judges 0 of 361 follow, lose at most 9.0 pp; fine-tuned judges at most 8.4\% follow, yet lose 31--99 pp. & The classifier misses none; no zero-shot judge finds the boundary rule's one miss. & --- \\
\addlinespace
AFTraj & Current-turn TF--IDF and GPT-4.1 both 82/163. & --- & Against current-turn TF--IDF: GPT-4.1 gains 46 and loses 46 (residual accuracy 56.8\%). & TF--IDF vs.\ GPT-4.1: $+17.6$ pp on injected, $-34.5$ pp on natural. Probe and DeepSeek V4 Pro swap ranks at 63.0\% injected. \\
\addlinespace
Aegis & All speakers 49.2\% micro-F1, GPT-4.1 50.0\%. & --- & Against all speakers: GPT-4.1 gains 124 exact sets and loses 9. & Gap to all speakers: 0.8 pp on micro-F1 (interval includes 0), 19.2 pp on exact sets. \\
\addlinespace
Video & Final-step guess 63.3\%, GPT-4.1 59.8\% (mean of DVD and EVA). & --- & Against the final-step guess (DVD): GPT-4.1 gains 27, loses 39. & --- \\
\bottomrule
\end{tabular}\par
\end{table*}

%% file: latex/related_work.tex
\section{Related Work}
\label{sec:related}
\paragraph{Input restriction and cue following.}
Partial-input baselines, which expose shortcut learning \citep{geirhos2020shortcut}, predict NLI labels from the hypothesis alone \citep{gururangan2018annotation,poliak2018hypothesis} and reading-comprehension answers from the question, the passage, or one sentence \citep{kaushik2018reading,chen2016cnn}. A successful partial-input baseline shows that a cue exists, not that models use it \citep{feng2019misleading}. Whether a model uses a cue is usually tested on new examples that break it \citep{niven2019probing,mccoy2019right} or on examples from unseen annotators, whose writing lacks the training annotators' cues \citep{geva2019annotator}. Cue following instead edits the same trajectory, moving one cue but not the error, so any change in the answer comes from the cue alone; measuring accuracy also catches judges that depend on a cue without following it.

\paragraph{Residual accuracy and aggregation.}
Hard-subset analyses score models on questions a heuristic cannot answer \citep{sugawara2018makes}. Residual accuracy reflects only gains, but the judge's accuracy minus the predictor's equals gains minus losses (Equation~\ref{eq:gainloss}), so we also report losses. Test populations and aggregation rules can decide rankings \citep{binette2024estimands,zhang2024aggregation}, and benchmark validity requires rankings that support claims about the intended capability \citep{bowman2021fix,bean2025construct,freiesleben2025epistemology}. We add the failure-mix weight at which two methods swap ranks and its distance from the released mixture.

\paragraph{Attribution benchmarks and judges.}
We audit Who\&When Pro \citep{liu2026whowhenpro}, which succeeds Who\&When \citep{zhang2025whowhen}, as well as AFTraj \citep{zhang2026aftraj} and Aegis \citep{kong2026aegis}. Other benchmarks allow several valid attributions per failure \citep{in2026mpbench} or log each agent's full inputs \citep{chen2026traceelephant}, and DoVer checks an attribution by editing the blamed step and rerunning \citep{ma2025dover}. Closest to our method is the shortcut test in CatchBench \citep{zhao2026catchbench}, a benchmark for predicting agent failures from the configuration declared before a run. A rule that ignores the configuration's content reaches perfect F1 on one of its six data sources, and its authors argue that a score cannot be interpreted until the labeling process has been tested for such shortcuts. We go further by testing whether judges use such cues, and find that fine-tuning on benchmark labels teaches them to. LLM judges of agent trajectories accept failed runs that end in confident language \citep{advani2026false} and raise many false alarms that recall alone hides \citep{mohammadi2026trajectory}. Unlike these studies, which ask whether a run failed, we target cues that point to the faulty step.

%% file: latex/conclusion.tex
\section{Conclusion}
\label{sec:conclusion}
On all three failure-attribution benchmarks we audit, a predictor without a cross-step check matches or beats GPT-4.1. On Who\&When Pro, one step's text gives away the labels in both systems that record reasoning, and judges fine-tuned on ALFAgent beat GPT-4.1 by 41 points but lose 31 to 99 when task-related reasoning is added after the error; on AFTraj, a classifier that sees one turn ties GPT-4.1 only at the released failure mix; on Aegis, micro-F1 hides GPT-4.1's 19-point exact-set lead over blaming every agent. A high score alone therefore cannot show that a judge checks across steps, which motivates our four tests. The tests need no new annotation, so we recommend that benchmark developers report the applicable tests next to the main score, using Table~\ref{tab:audit-summary} as a template.

%% file: latex/limitations.tex
\section*{Limitations}

We run cue following only on ALFAgent, where one step's text recovers all 311 labels and all errors come from one injection pipeline. The judges we run on ALFAgent answer our own prompt, not the benchmark's (Appendix~\ref{app:official-pro}). Our fine-tuned judges cover one subset, two model sizes, and two seeds each; their Task continued losses hold across seeds, but their Template continued losses vary widely (Table~\ref{tab:finetuned}). We use three zero-shot judges on ALFAgent but only GPT-4.1 on video and Aegis, and we evaluate Aegis on one release of its data (Appendix~\ref{app:aegis-versions}).

Some results have other possible explanations. On ALFAgent, the error-step text that the classifier picks up may state the mistake plainly enough for a careful reader to spot it from that text alone, or it may carry traces of how the error was injected; our tests do not separate these two cases. The paraphrase test changes only the wording of the error step and keeps its claims, so it cannot show whether judges rely on those claims. An automatic check that the paraphrases keep their meaning (Appendix~\ref{app:paraphrase}) used GPT-4.1, which is also a judge. The 60 human-reviewed pairs do not depend on this check. On AFTraj, every coding failure is injected, so part of the injected--natural difference may reflect domain. Within the first-alarm test's 68 agentic failures, the injected-minus-natural gap keeps its sign for five of six methods, but every interval for this gap includes zero. The separate experiment on 468 agentic failures shows that injected errors carry recognizable text within the agentic domain too.

Finally, we score only against the released annotations, so a judge that blames a different but also valid step or agent counts as wrong.

\section*{Ethical Considerations}

This work audits public benchmarks; it collects no data from people and releases no new model or dataset. Its main risk is that the cues it documents could be used to inflate a judge's score. The four tests are designed to expose such inflation, and reporting them next to the main score makes it visible. We use the three benchmarks under their licenses (Appendix~\ref{app:repro}). The authors themselves carried out the human checks of generated text (Appendices~\ref{app:task-related-continuation} and~\ref{app:paraphrase}). All experiments together used about 20 GPU-hours (Appendix~\ref{app:repro}).

%% file: latex/appendix/evaluation.tex
\section{Evaluation Details}
\label{app:repro}

\paragraph{Task groups.}
Records that come from the same task form a \emph{task group}. For Aegis, conversations built from the same user query form a \emph{query group}. Whenever we split or resample data, a group stays together.

\paragraph{Data.}
Who\&When Pro text has 6,257 trajectories from eight agent systems. The boundary rule (Appendix~\ref{app:text-fields}) and the position rules, which always predict the final or the penultimate step (Appendix~\ref{app:text-position}), are evaluated on all of them. The independent-step classifier on smolagents uses that system's 3,276 trajectories, which form 1,411 task groups (Appendix~\ref{app:smolagents}). Every other text experiment uses the 311 ALFAgent trajectories (119 task groups, 30 steps each).

Some ALFAgent tests use fewer judges. Among the three zero-shot judges, Task continued and the paraphrase test (Section~\ref{sec:results}) are run with GPT-4.1 and Claude Sonnet 5 only, without o3. A control version of Template continued, called Template before error, appends the same sentence to every step before the error, which keeps the reasoning boundary at the error. Template before error is run on three models only, the two second-seed fine-tuned models and Qwen3-8B before fine-tuning (Table~\ref{tab:finetuned}), and not on GPT-4.1, Claude Sonnet 5, or o3. These three Qwen models are also the only Qwen models that run the paraphrase test.

The paraphrase test uses 285 of the 311 trajectories, because it needs a step with reasoning before the error and the 26 trajectories whose error is at step 1 have none. One of the 26 has no reasoning at any step, so it is also the one trajectory that the boundary rule misses.

Who\&When Pro video has 168 DVD and 748 EVA trajectories. Video results that combine the two systems take the mean of their accuracies, so DVD and EVA count equally. AFTraj uses the 163 failures in the first-alarm test. Aegis uses 600 conversations in 441 query groups. Step and turn numbers follow each benchmark's released numbering; AFTraj turns start at 0.

\paragraph{Choosing restricted-input predictors.}
\textbf{A restricted-input predictor must not be chosen on the records it is scored on.} The Aegis all-speakers predictor and the AFTraj fixed-turn predictor are chosen from candidates on one half of the groups and scored on the other half, and then the halves are swapped. We sort the group identifiers and cut the sorted list in the middle, so the split is reproducible. The choice maximizes micro-F1 for Aegis and exact accuracy for AFTraj. Current-turn TF--IDF is trained on AFTraj tasks outside the first-alarm test (Appendix~\ref{app:aftraj-setup}). The Who\&When Pro position rules, the video position predictor, and the independent-step classifiers on ALFAgent and smolagents are chosen or trained by five-fold cross-validation over task groups. The fine-tuned judges are trained on the same five folds as the ALFAgent classifier. The boundary rule (predict the last step with nonempty reasoning) has no parameter, so nothing is chosen. Appendices~\ref{app:alfagent}--\ref{app:video} list the candidate predictors.

\paragraph{Intervals and invalid answers.}
Every interval comes from resampling task groups (query groups for Aegis) 10,000 times. For a difference between two methods, both methods are scored on the same resample. The bootstrap seed is 0, except for video and for the 468-failure experiment on AFTraj (Appendix~\ref{app:current-turn}), which use seed 1. An answer that cannot be scored counts as a miss. Every model returns a scorable answer on at least 95\% of records in every experiment.

\paragraph{Models.}
The model identifiers are \nolinkurl{openai/gpt-4.1}, \nolinkurl{anthropic/claude-sonnet-5}, \nolinkurl{openai/o3} (medium reasoning effort), \nolinkurl{Qwen/Qwen2.5-1.5B-Instruct}, \nolinkurl{Qwen/Qwen3-8B}, \nolinkurl{Qwen/Qwen3-32B}, and \nolinkurl{deepseek/deepseek-v4-pro}. DeepSeek V4 Pro has two roles. It is one of the AFTraj judges, and it writes the reasoning added in Task continued and the paraphrases used in the paraphrase test. The writing calls were served by snapshot \nolinkurl{deepseek-v4-pro-0813}.

\paragraph{Licenses.}
Who\&When Pro is released under CC BY 4.0, AFTraj under the MIT License, and Aegis under the MIT License. We use the three benchmarks only to evaluate failure-attribution methods, which these licenses permit, and we do not redistribute their data.

\paragraph{Compute and software.}
Qwen2.5-1.5B, Qwen3-8B, and Qwen3-32B have about 1.5, 8, and 32 billion parameters. GPT-4.1, Claude Sonnet 5, o3, and DeepSeek V4 Pro are called through their providers' APIs. Fine-tuning and evaluating the Qwen judges of Appendix~\ref{app:finetuned} took about 8 GPU-hours, with each run on one NVIDIA RTX A6000 GPU (48 GB). Extracting the Probe features and running the two Qwen3 judges on AFTraj took about 12 more GPU-hours. The TF--IDF classifiers run on CPUs. The fine-tuned judges use PyTorch 2.11.0, Transformers 5.19.0, and PEFT 0.21.2, and the independent-step classifiers use scikit-learn 1.7.2.

\paragraph{AI assistance.}
The authors used generative AI for analysis code, figures, and editing, and take responsibility for the interpretation and the submission.

%% file: latex/appendix/alfagent.tex
\section{Who\&When Pro Text: Protocols and Complete Results}
\label{app:alfagent}

This appendix gives the details behind Section~\ref{sec:alfagent}. Appendices~\ref{app:text-fields} and~\ref{app:text-position} compare ALFAgent with the other Who\&When Pro text systems. Appendix~\ref{app:official-pro} describes how the judges are prompted and what changes under the benchmark's own prompt. Appendix~\ref{app:independent-step} describes the independent-step classifier and checks that its accuracy does not come from recurring phrases or from reasoning length. Appendix~\ref{app:reasoning-strata} shows, with three examples, what the agent's reasoning says at the error step. Appendices~\ref{app:task-related-continuation} and~\ref{app:paraphrase} give the construction and full results of the two interventions the main text summarizes, Task continued and the paraphrase test. Appendix~\ref{app:smolagents} runs the classifier on smolagents, and Appendix~\ref{app:finetuned} fine-tunes two judges on the labels.

\subsection{Reasoning boundary across text systems}
\label{app:text-fields}

\begin{table}[!t]
\centering\footnotesize
\setlength{\tabcolsep}{2.5pt}
\caption{The boundary rule and the position rule on each Who\&When Pro text system. The boundary rule predicts the last step with nonempty reasoning; the position rule (Appendix~\ref{app:text-position}) predicts the final or penultimate step, chosen per system. \emph{Reasoning}: trajectories with any nonempty reasoning. \emph{Boundary} and \emph{Position}: the number of trajectories on which each rule is right, with the percentage of the system's trajectories in parentheses. ``---'': the system has no per-step reasoning, so the boundary rule makes no prediction.}
\label{tab:text-boundary-app}
\begin{tabular}{@{}lrrrr@{}}
\toprule
System & $N$ & Reasoning & Boundary & Position \\
\midrule
ALFAgent & 311 & 310 & 310 (99.7) & 2 (0.6) \\
smolagents & 3,276 & 3,276 & 1,707 (52.1) & 1,707 (52.1) \\
MathChat & 920 & 0 & --- & 725 (78.8) \\
Magentic-One & 811 & 0 & --- & 454 (56.0) \\
Debate & 445 & 0 & --- & 204 (45.8) \\
MacNet & 261 & 0 & --- & 109 (41.8) \\
DyLAN & 152 & 0 & --- & 0 (0.0) \\
MetaGPT & 81 & 0 & --- & 14 (17.3) \\
\bottomrule
\end{tabular}
\end{table}

\textbf{The boundary rule tracks the error only in ALFAgent} (Table~\ref{tab:text-boundary-app}). There it is right on 310 of 311 trajectories, because the reasoning stops before the final step, exactly at the error. No other system shows this.

In smolagents the rule amounts to guessing the final step. The reasoning field is nonempty through the last step in all but nine of the 3,276 trajectories, so the rule predicts the final step there. Its 1,707 hits are exactly the trajectories whose error is at the final step. In the nine trajectories where reasoning stops earlier, the rule is wrong.

The other six systems have no per-step reasoning field, so the rule makes no prediction, which is scored as a miss. Magentic-One has 1,358 fields named ``reasoning,'' but they are arguments of tool calls rather than a per-step reasoning field, so we do not count them.

\subsection{Position rules across text systems}
\label{app:text-position}

\textbf{Position, a strong cue in most other Who\&When Pro text systems, is not one in ALFAgent.} A position rule ignores the text and always predicts the same position. We allow two positions, the final step and the penultimate step, and choose one for each system by five-fold cross-validation over task groups. Every fold chooses penultimate for ALFAgent, MacNet, and MetaGPT, and final for the other five systems. The resulting rules are right on 3,215 of all 6,257 text trajectories (51.4\%) and on 78.8\% of MathChat trajectories, but on only 2 of the 311 ALFAgent trajectories.

Three scoring details affect these numbers. For Debate and DyLAN the annotation is a whole round, a block of consecutive steps, instead of a single step. A prediction therefore counts as correct if its step falls anywhere in that round. Some MacNet labels also list alternative steps, and the benchmark's scorer counts a prediction at any of them as correct. Two Magentic-One trajectories have a single step and therefore no penultimate step; when cross-validation compares the two positions, the penultimate rule is scored as wrong on both. Table~\ref{tab:text-boundary-app} gives the result for each system; 95\% intervals from resampling task groups are in the supplementary material.

\subsection{Judge prompts and settings}
\label{app:official-pro}

All our ALFAgent judges, zero-shot and fine-tuned, receive the same simple prompt, which we call the \emph{controlled prompt}; only in one experiment do GPT-4.1 and Claude Sonnet 5 also answer the benchmark's own prompt (Table~\ref{tab:alfagent-official}B). The controlled prompt shows the 30 numbered steps of one trajectory and asks only for the step of the injected error, as \texttt{\{"step": <integer>\}}. The gold label is never in the prompt.

GPT-4.1 and Claude Sonnet 5 run at temperature 0, and o3 runs at medium reasoning effort. o3 produces 15 invalid outputs out of 311 under No reasoning. Every other combination of judge and input version produces at most three.

\textbf{The controlled prompt does not understate the judges}: under the benchmark's own prompt, their step accuracy is lower, 50.8\% against 58.8\% for GPT-4.1 and 51.4\% against 57.2\% for Claude Sonnet 5. That prompt asks, in a single response, for the agent that made the error, the step, the error mode (one of 17 in the benchmark's taxonomy), and a one-line reason. We score the responses with the parser released with the benchmark (Table~\ref{tab:alfagent-official}B, where Joint means that agent, step, and mode are all right). Both judges almost always name the agent correctly: GPT-4.1 in 100.0\% of trajectories and Claude Sonnet 5 in 97.4\%. Getting agent, step, and mode all right is much rarer: 25.1\% for GPT-4.1 and 22.8\% for Claude Sonnet 5.

Template continued raises o3's accuracy by a net 4.5 points (Table~\ref{tab:alfagent-arms}). Under Template continued, o3 gives a different answer than under Original on 88 of the 311 trajectories. Of these 88 changes, 33 go from wrong to right, 19 go from right to wrong, and 36 go from one wrong step to another.

When GPT-4.1 and Claude Sonnet 5 are wrong under Original, they mostly name a step after the error (Table~\ref{tab:alfagent-errors}). GPT-4.1 places 87 of its 128 wrong predictions after the error, on steps without reasoning, and Claude Sonnet 5 places 105 of its 133 there. These late predictions are a median of 8 steps after the error.

\subsection{Independent-step classifiers}
\label{app:independent-step}

Section~\ref{sec:alfagent} uses a classifier that scores each step on its own. This subsection gives its settings, shows which parts of a step it needs, and lists the words it relies on.

The classifier is a logistic regression that labels each step as the error step or not, with class weights balanced between the two labels. It uses 20,000 unigram and bigram TF--IDF features with sublinear term frequency and at most 2,000 training iterations. A step is represented by its own fields only, with the field names removed. For a held-out trajectory, the prediction is the highest-scoring step, and ties go to the earliest step. We use five folds in which all trajectories of a task group (Appendix~\ref{app:repro}) fall in the same fold. To build them, we sort the task groups, shuffle them with seed 0, and assign them to the five folds in turn.

\begin{table}[!t]
\centering\footnotesize
\setlength{\tabcolsep}{4pt}
\caption{What predicts the annotated error step in ALFAgent. A: accuracy (five folds, 311 records) of the independent-step classifier when it sees different parts of a step; the last two rows instead show small models that see only the length of the reasoning or whether it is present. B: the words or word pairs with the ten largest positive weights $\beta$ and those with the ten most negative weights, from one fit of the classifier to the full step text of all 311 records. A positive weight makes a step more likely to be chosen as the error. The row ``repetitive loop'' merges six overlapping words and word pairs that receive the same weight: \emph{repetitive}, \emph{loop}, \emph{repeating}, \emph{repetitive loop}, \emph{loop repeating}, and \emph{repeating go}.}
\label{tab:alfagent-local-app}
\label{tab:alf-length-app}
\label{tab:alf-lexical-app}
\textbf{A. Restricted inputs}\par\smallskip
\begin{tabular}{@{}lr@{}}
\toprule
Available information & Acc. (\%) \\
\midrule
All step text & 100.0 \\
Reasoning only & 100.0 \\
All step text, recurring phrases deleted & 98.4 \\
All text except reasoning & 21.2 \\
\quad Observation only & 20.6 \\
\quad Action only & 11.6 \\
Reasoning length (best of four models) & 28.9 \\
Whether reasoning is present & 8.4 \\
\bottomrule
\end{tabular}\par
\medskip
\textbf{B. Highest-weight words}\par\smallskip
\begin{tabular}{@{}lrlr@{}}
\toprule
Positive & $\beta$ & Negative & $\beta$ \\
\midrule
already & $+4.31$ & you & $-3.73$ \\
can & $+4.11$ & examine & $-3.45$ \\
keeps & $+4.06$ & move & $-2.45$ \\
so & $+4.01$ & not & $-2.45$ \\
now & $+3.85$ & see & $-2.38$ \\
before & $+3.70$ & closed & $-2.18$ \\
item & $+3.66$ & logical & $-2.17$ \\
repetitive loop (6 n-grams) & $+3.58$ & open & $-2.10$ \\
times & $+3.57$ & you see & $-2.09$ \\
that & $+3.38$ & but & $-2.01$ \\
\bottomrule
\end{tabular}
\end{table}

\textbf{The classifier needs only the reasoning}: with the full step text, or with the reasoning alone, it finds every error, and without the reasoning it finds 21.2\% (Table~\ref{tab:alfagent-local-app}A). Two simpler explanations of why the reasoning alone suffices can be ruled out.

First, the classifier does not need recurring phrases such as ``Repetitive loop''. After lowercasing the text, we delete every span that begins with ``repetitive loop'', ``no prior actions to repeat'', ``earlier i saw'', or ``earlier i noticed'' and runs to the next period. Retrained on the edited text, the classifier still finds the annotated step in 306 of the 311 held-out trajectories (98.4\%).

Second, the length of the reasoning is not enough. A classifier that sees only the length of the reasoning, with or without the step's position, reaches 28.9\% at best. That is the best of four small models: two inputs (length alone, or length and position) and two classifiers (logistic regression, or a depth-three tree). Length is $\log(1+\text{characters})$ of the trimmed reasoning, and position is $(t-1)/(T-1)$ for step $t$ of $T$.

Whether a step has reasoning is a cue only when steps are compared. A classifier that knows only whether a step has reasoning finds 8.4\%, far below the boundary rule's 99.7\%. This classifier sees each step on its own, so it cannot tell the last step with reasoning from the earlier steps with reasoning. With ties going to the earliest step, it picks step 1, which is the error in 26 of the 311 trajectories. The boundary rule finds the last step with reasoning by comparing steps.

The classifier relies on words that declare progress or a loop (Table~\ref{tab:alfagent-local-app}B). For this table the classifier is fit once to the full step text of all 311 records. The five fits from the cross-validation folds have largely the same top words. Large positive weights go to such words, for example \emph{already}, \emph{now}, \emph{keeps}, \emph{times}, and \emph{repetitive loop}. Large negative weights go to words of the environment's observations and of searching, such as \emph{you} (as in ``You see''), \emph{examine}, \emph{see}, and \emph{open}. Appendix~\ref{app:paraphrase} tests how far this signal survives rewording.

\subsection{The independent-step classifier on smolagents}
\label{app:smolagents}
\input{latex/table/smolagents_local.tex}
\textbf{One step's text is a cue in smolagents too}: from the reasoning alone, the same independent-step classifier picks the annotated step in 95.2\% $[94.3,96.0]$ of trajectories, and in 90.8\% of the 1,569 whose error is not at the final step (Table~\ref{tab:smolagents-local}). smolagents is the largest Who\&When Pro text system and the only other one with a per-step reasoning field. Its reasoning runs to the final step in all but 9 trajectories, so the boundary rule there reduces to guessing the final step (52.1\%).

The signal lies in the wording of the reasoning, not in its form. The error step's reasoning never puts code in a Markdown \texttt{python} block (0 of 3,276 error steps), while 3,593 of the 8,013 steps before the error and 1,348 of the 3,649 steps after it do. Removing every code block and code marker from the reasoning erases this difference in form, yet it raises accuracy from 95.2\% to 96.2\%.

Adding the other fields to the reasoning only lowers accuracy. Adding the code field or the observation field to the reasoning lowers accuracy to 93.9\% and 92.5\%, and adding both, which gives the full step text, lowers it to 91.9\%. Without the reasoning, the classifier finds 48.5\% of the errors that are not at the final step, against 90.8\% with the reasoning alone.

\subsection{What the error-step reasoning says}
\label{app:reasoning-strata}
\label{app:reasoning-pairs}

This subsection returns to ALFAgent. At the annotated error step, the agent's reasoning follows a template in two of the six error modes of the benchmark's taxonomy and is free text in the other four. We read ten records of each mode closely. The counts below cover all records of a mode. 310 of the 311 error steps have reasoning.

In \emph{looping behavior} (53 records) the reasoning refers to repetition outright: 46 begin ``Repetitive loop'', and 7, all with the error at step 1, where no earlier step exists, begin ``No prior actions to repeat''. In \emph{hallucination} (42 records) it claims a memory, such as ``Earlier I saw\ldots'', which can be checked only against earlier observations.

In the other four modes the reasoning is free text. In \emph{reasoning error} it argues for substituting the wrong object. In \emph{inadequate verification} it assumes that an unfinished step was done. In \emph{task misunderstanding} and \emph{context loss} it looks plausible on its own.

\textbf{The judges do worse, not better, when the error step states a loop outright} (Table~\ref{tab:reasoning-strata}A). We call a record a \emph{loop record} if the reasoning at its error step contains ``repetitive loop'' or ``no prior actions to repeat'', ignoring case; these are the 53 looping-behavior records. Under Original, GPT-4.1 and Claude Sonnet 5 both do much worse on the 53 loop records than on the other 257 records with reasoning. GPT-4.1 is right on 8 of the 53 (15\%) and on 175 of the 257 (68\%). Claude Sonnet 5 is right on 10 of the 53 (19\%) and on 168 of the 257 (65\%). The judges need the reasoning (Section~\ref{sec:alfagent}), but what they use in it is not a recurring phrase that announces the error.

\input{latex/table/reasoning_strata.tex}

In cross-validation, the independent-step classifier selects the annotated step in all three records below. For each, we give the complete reasoning at the annotated error and at the step before it. Records 0012 and 0023 are the two examples in Figure~\ref{fig:alfagent-examples}; record 0001 is the first loop record in ID order. IDs carry the prefix \texttt{alfworld\_alfagent\_}.

\paragraph{Record 0012, reasoning error.}
\emph{Preceding step (5):} ``The cellphone is not found in drawer 2 based on the visible items, which only include keychains. Since the cellphone has not been found yet, the next step is to check other furniture where the cellphone might be, such as the sidetable 1. Going to sidetable 1 will allow us to check that location for the cellphone.''

\emph{Annotated error (6):} ``A cellphone is functionally the same kind of small handheld media device as a CD, so this task can be satisfied by placing the matching portable item into the desk. Since the sidetable already has a CD available, I should pick it up now and use it as the target object for the desk placement.''

\paragraph{Record 0023, inadequate verification.}
\emph{Preceding step (7):} ``Drawer 3 is open but empty, so the book is not here either. The task requires finding a book, so I should continue searching other locations. The remaining drawers and other furniture are possible places to check, so moving next to drawer 4 is the logical step to continue the systematic search for the book.''

\emph{Annotated error (8):} ``The drawers have already been checked, and the book has been accounted for, so the next step is to complete the placement. Going to the sofa now puts me in position to put the book where it belongs without wasting more time searching.''

\paragraph{Record 0001, looping behavior.}
\emph{Preceding step (5):} ``To put the pencils in drawer 1, I first need to open drawer 1 since it is currently closed. Opening the drawer is the necessary next step before I can put the pencils inside.''

\emph{Annotated error (6):} ``Repetitive loop: repeating `go to drawer 1' 3 times''

\subsection{Task continued}
\label{app:task-related-continuation}

Task continued checks whether text about the task, which is harder to ignore than a generic sentence, pulls the judges to a moved boundary. Template continued (Section~\ref{sec:results}) adds one generic sentence to each empty reasoning field after the error, so that the last step with reasoning becomes step 30, and almost no prediction follows the moved boundary. Task continued moves the boundary in the same way but adds text about the task itself.

Task continued fills every empty reasoning field after the error. There are 6,540 such fields across the 311 trajectories. Reasoning then runs through step 30 in every trajectory. An automatic check confirms that the original actions, observations, and reasoning are unchanged, so the error stays at its step.

The added passages contain no error terms, and 71 of 72 pass a human review. DeepSeek V4 Pro writes each added passage as one first-person sentence that gives a purpose for the action at that step. For the passage at step $k$, the generator sees the task and every step before the error in full. It also sees the actions and observations from the error step through step $k$, but not the reasoning at the error step. It is not told which step is the annotated error, and it is not given the error mode, any judge's answer, any step after step $k$, or the passages written for other fields. An automatic check confirms that each of the 6,540 passages is a single line of 12 to 55 words and contains no word from a list of error terms, such as \emph{error}, \emph{wrong}, \emph{fail}, or \emph{loop}. A human review covers 72 passages. The sample has four passages for each combination of the six error modes (Appendix~\ref{app:reasoning-strata}) and three ranges of the passage's step (steps 1--10, 11--20, and 21--30). A passage passes if it gives a concrete purpose for its action, agrees with the action and observation at its step, and does not comment on errors. Of the 72 passages, 71 pass, and the remaining one has a minor wording issue about the agent's plan.

\input{latex/table/alfagent_arms.tex}

Task continued does not pull GPT-4.1 or Claude Sonnet 5 to the moved boundary (Table~\ref{tab:alfagent-arms}). The old boundary is the last step with reasoning under Original. Under Original, the two judges make 361 correct predictions at the old boundary (183 for GPT-4.1 and 178 for Claude Sonnet 5). Under Task continued, none of these 361 predictions moves to step 30, and 294 stay at the old boundary (155 for GPT-4.1 and 139 for Claude Sonnet 5). Of the 67 predictions that leave the old boundary, 52 move past the error (25 for GPT-4.1 and 27 for Claude Sonnet 5), and 14 move before it (3 for GPT-4.1 and 11 for Claude Sonnet 5). The remaining one becomes an invalid answer from Claude Sonnet 5. Most of these predictions move past the error, where most wrong predictions already fall under Original (Table~\ref{tab:alfagent-errors}). \textbf{The added text moves some predictions away from the error, but it moves none of them to step 30.} Across all 311 trajectories under Task continued, three answers from Claude Sonnet 5 are invalid, including the one above, because they do not give a step number in the required format; an invalid answer counts as a miss (Appendix~\ref{app:official-pro}).

Task continued costs the zero-shot judges more than Template continued does, but at most 9.0 points. Relative to Original, accuracy falls by 4.2 points for GPT-4.1 and by 9.0 points for Claude Sonnet 5 (Table~\ref{tab:alfagent-arms}). The 95\% task-group bootstrap interval of the GPT-4.1 change, $[-8.4, 0.0]$, includes zero. Under Task continued, each judge is right on 20 fewer trajectories (6.4 points) than under Template continued. GPT-4.1 drops from 190 correct trajectories (61.1\%) to 170 (54.7\%), and Claude Sonnet 5 drops from 170 correct trajectories (54.7\%) to 150 (48.2\%).

Task continued costs the independent-step classifier, which scores each step on its own and so cannot see where the reasoning stops, 9.0 points, far below the fine-tuned judges' losses of 31 to 99 points (Appendix~\ref{app:finetuned}). We use the classifier's reasoning-only version (Appendix~\ref{app:independent-step}), trained on Original. Under Task continued, it still selects the annotated step in 283 of the 311 trajectories (91.0\%). Its other 28 predictions move to later steps, all of which carry added text.

The zero-shot judges lose few trajectories and gain some, and most trajectories they lose are not among those the classifier loses, so their losses mostly do not come from the added text that misleads the classifier; each fine-tuned run, in contrast, loses many and gains almost none. A judge loses a trajectory when its answer is right under Original and wrong under Task continued. It gains a trajectory in the opposite case. Of the 28 trajectories that GPT-4.1 loses, only 3 are among the 28 that the classifier loses. Of the 39 trajectories that Claude Sonnet 5 loses, only 6 are among the classifier's 28. The zero-shot judges also gain trajectories. GPT-4.1 gains 15, and Claude Sonnet 5 gains 11. In contrast, each fine-tuned run loses between 96 and 309 trajectories, and the four runs together gain one (Appendix~\ref{app:finetuned}).

\subsection{Paraphrase test}
\label{app:paraphrase}

\textbf{Judge accuracy does not change detectably when we rewrite the reasoning at the error step} (Gold paraphrased) or, as a control, at the step before it (Prior paraphrased). The paraphrase test therefore finds no sign that the judges depend on the exact wording of the reasoning.

The test leaves out the 26 records whose error is at step 1, because they have no earlier step, and uses the other 285. In each of the two conditions, DeepSeek V4 Pro rewrites one reasoning string per record, so it rewrites 570 strings in total. It sees only the string it rewrites. The rewrites are slightly longer than the originals. Under Gold paraphrased, the average string grows from 39.5 to 40.9 words. Under Prior paraphrased, it grows from 55.1 to 56.9 words.

The rewrites keep the meaning. A human review covers 60 pairs, five from each condition for each of the six error modes. It rates 56 pairs as equivalent in meaning, 4 as differing in a minor way, and none as differing in a material way. GPT-4.1 also rates 569 of the 570 pairs as equivalent, but it is one of the judges, so this check is not independent. The exception is record 0001 under Gold paraphrased, where ``repeating `go to drawer 1' 3 times'' became ``say `go to drawer 1' 3 times''. We keep the pair as generated.

\begin{table}[!t]
\centering\footnotesize
\setlength{\tabcolsep}{4pt}
\caption{Paraphrase results on the 285 records whose error is not at step 1. A: number correct (accuracy, \%) for the independent-step classifier on all step text and for the two judges, then paired changes in points with 95\% task-group bootstrap intervals. Gold $-$ Prior compares the two paraphrase conditions. B: for each judge and condition, how many predictions stay on the same step as under Original, how many turn from right to wrong or from wrong to right, and how many answers are valid (in the required format). ---: not computed.}
\label{tab:paraphrase-results}
\label{tab:paraphrase-records}
\textbf{A. Accuracy and paired changes}\par\smallskip
\resizebox{\ifdim\width>\columnwidth\columnwidth\else\width\fi}{!}{%
\begin{tabular}{@{}lrrr@{}}
\toprule
 & Classifier & GPT-4.1 & Claude Sonnet 5 \\
\midrule
Original & 285 (100.0) & 174 (61.1) & 167 (58.6) \\
Gold paraphrased & 273 (95.8) & 178 (62.5) & 168 (58.9) \\
Prior paraphrased & 282 (98.9) & 169 (59.3) & 166 (58.2) \\
Gold $-$ Original & --- & \scoreci{$+1.40$}{$-2.15$}{$+4.91$} & \scoreci{$+0.35$}{$-2.70$}{$+3.50$} \\
Prior $-$ Original & --- & \scoreci{$-1.75$}{$-5.45$}{$+1.77$} & \scoreci{$-0.35$}{$-3.64$}{$+3.00$} \\
Gold $-$ Prior & --- & \scoreci{$+3.16$}{$-0.36$}{$+7.00$} & \scoreci{$+0.70$}{$-2.54$}{$+3.91$} \\
\bottomrule
\end{tabular}}\par
\medskip
\textbf{B. Per-record changes}\par\smallskip
\resizebox{\ifdim\width>\columnwidth\columnwidth\else\width\fi}{!}{%
\begin{tabular}{@{}llrrrr@{}}
\toprule
Judge & Condition & \shortstack[r]{Same\\step} & \shortstack[r]{Right\\$\to$ wrong} & \shortstack[r]{Wrong\\$\to$ right} & Valid \\
\midrule
GPT-4.1 & Gold & 241 & 12 & 16 & 285/285 \\
GPT-4.1 & Prior & 248 & 15 & 10 & 285/285 \\
Claude Sonnet 5 & Gold & 239 & 10 & 11 & 282/285 \\
Claude Sonnet 5 & Prior & 239 & 11 & 10 & 283/285 \\
\bottomrule
\end{tabular}}
\end{table}

Neither rewrite changes either judge's accuracy detectably: every interval in Table~\ref{tab:paraphrase-results}A includes zero. Under Gold paraphrased, GPT-4.1 gains 1.40 points and Claude Sonnet 5 gains 0.35 points. Table~\ref{tab:paraphrase-results}B counts individual predictions. For each judge and condition, at least 239 of the 285 predictions stay on the same step as under Original, and about as many turn from right to wrong as from wrong to right.

The rewrites weaken the wording cue without removing it. To see this, we apply the independent-step classifier of Appendix~\ref{app:independent-step} without retraining it on the rewrites. This version reads the full text of each step, so it sees the unchanged actions and observations as well as the reasoning. It still finds the annotated step in 273 of the 285 records under Gold paraphrased and in 282 of the 285 under Prior paraphrased. For each record, we also compute the classifier's margin. The margin is the decision value of the annotated step minus the highest decision value of any other step in the same trajectory, so a positive margin means that the classifier picks the annotated step. Under Original, the mean margin is 3.67. Under Gold paraphrased, the margin falls by 1.55 on average (95\% task-group bootstrap interval $[1.41,1.68]$). Under Prior paraphrased, it falls by 0.27 on average ($[0.21,0.33]$). Under Prior paraphrased, the classifier misses 3 records, and in all 3 it picks the rewritten step just before the error. Under Original, it misses none of the 285 records.

\subsection{Fine-tuned judges}
\label{app:finetuned}

\input{latex/table/finetuned_judge.tex}

The judges of Section~\ref{sec:results} are zero-shot. To see what a judge learns from the labels, we fine-tune Qwen2.5-1.5B-Instruct \citep{qwen2024qwen25} and Qwen3-8B \citep{qwen2025qwen3} with LoRA \citep{hu2022lora}. We call them the 1.5B model and the 8B model. The 8B model runs in non-thinking mode, in which it answers without first generating a reasoning trace. Each model is trained to output the annotated step when given the prompt that the zero-shot judges receive (the controlled prompt; Appendix~\ref{app:official-pro}) with the Original input. We use the five task-disjoint folds of Appendix~\ref{app:independent-step}. Each record is answered by the model trained on the other four folds, so the model never sees the record's task group in training. LoRA has rank 16 (alpha 32, dropout 0.05) on all attention and MLP projections. Training uses AdamW with an effective batch size of 4 and computes the loss on the answer tokens only. It runs for three epochs at learning rate $2\times10^{-4}$ with 5\% warmup and linear decay. We train each model twice, once with seed 0 and once with seed 1, and decode answers greedily. We call the two seed-1 models the second-seed models (Seed 1 in Table~\ref{tab:finetuned}). On the second-seed models and on Qwen3-8B before fine-tuning, we also run three controls: Template before error, Gold paraphrased, and Prior paraphrased (Appendix~\ref{app:paraphrase}). Template before error appends the Template continued sentence to every step before the error, which keeps the reasoning boundary at the error.

Fine-tuning lifts both models to near 100\% (Table~\ref{tab:finetuned}). Before fine-tuning, the 1.5B model is near chance (5.5\%), and the 8B model scores 51.4\%, close to the zero-shot judges of Section~\ref{sec:results}. After fine-tuning, each of the four runs finds the annotated step in 310 or 311 of the 311 held-out records. Taken alone, these scores would suggest a judge that checks the error step against the other steps. The tuned models need the reasoning, since without it their accuracy falls to between 8.4\% and 15.8\%. Removing field names (Field-neutral) costs each run at most 3.5 points, so the field names matter little. The tuned models also depend on where the reasoning stops.

Task continued, which leaves the error step unchanged (Appendix~\ref{app:task-related-continuation}), costs the fine-tuned judges 31 to 99 points. Under Task continued, the 1.5B model falls to 2.3\% and 0.6\% (seeds 0 and 1), and the 8B model falls to 64.0\% and 68.8\%. The 8B model before fine-tuning loses only 4.5 points. Every answer that a tuned model loses moves to a step after the error, mostly toward the end of the trajectory. The 1.5B model loses 304 and 309 answers (seeds 0 and 1), and 216 and 175 of these land on step 29. The 8B model loses 111 and 96 answers (seeds 0 and 1), and in both seeds the median step they move to is step 25. Across the four runs, only one answer turns from wrong to right. In each run, at most 8.4\% of the answers at the old boundary (26 of 310) move to step 30. The tuned models therefore do not simply pick the last step with reasoning.

Template continued shows the same pattern. Its generic sentence costs the 1.5B model 29.6 and 59.2 points (seeds 0 and 1), and it costs the 8B model 13.2 and 1.3 points. Yet no answer of any fine-tuned run moves to step 30, so the follow rate $\mathcal F$ of Equation~\ref{eq:following} is 0 in every run. \textbf{A judge can therefore depend on a cue without following it.} The 8B model loses 41 and 4 answers (seeds 0 and 1), and every one of them moves to the step right after the error. That step is the first step after the original reasoning ends, so it is also the first step that carries the added sentence. These answers thus still track the end of the original reasoning, landing just past it, rather than the new last step with reasoning, which is step 30 ($c(\tau(X))$ in Equation~\ref{eq:following}). For comparison, the three zero-shot judges rarely follow the moved boundary either: 1 of their 536 correct answers at the old boundary moves to step 30.

The three controls change the accuracy of the second-seed models by less than 1 point. Template before error costs nothing ($+0.3$ points for the 1.5B model and 0.0 points for the 8B model), although it adds the same sentence as Template continued. Gold paraphrased costs nothing either (0.0 points for both models on the 285 records of the paraphrase test). Prior paraphrased changes accuracy by $+0.4$ points for the 1.5B model and by $-0.7$ points for the 8B model.

The fine-tuned judges' losses under Task continued do not come from words in the added passages that mark error steps; two results show this. First, the five words with the largest positive weights in the independent-step classifier (Table~\ref{tab:alfagent-local-app}B) are all rarer in the added passages than in the reasoning at the error steps. The table below gives, for each word, the percentage of the 6,540 added passages and of the 310 error-step reasoning strings that contain it. The error step of the remaining trajectory has no reasoning.
\begin{center}\footnotesize
\begin{tabular}{@{}lrr@{}}
\toprule
Word & Added passages (\%) & Error steps (\%) \\
\midrule
\emph{already} & 6.5 & 34.5 \\
\emph{can} & 29.6 & 32.3 \\
\emph{keeps} & 0.0 & 22.3 \\
\emph{so} & 34.7 & 56.1 \\
\emph{now} & 3.8 & 32.6 \\
\bottomrule
\end{tabular}
\end{center}
Second, the sentence of Template continued contains none of these words and no words about the task. It still costs the fine-tuned judges 1.3 to 59.2 points when it is added after the error, and it costs the second-seed models nothing when it is added before the error. We therefore find no sign that the fine-tuned judges respond to the wording of the added text.

%% file: latex/table/smolagents_local.tex
\begin{table}[!t]
\centering\footnotesize
\setlength{\tabcolsep}{4pt}
\caption{The independent-step classifier on smolagents, with the settings and fold procedure of Appendix~\ref{app:independent-step} (five folds over 1,411 task groups). Exact accuracy (\%) on all 3,276 records and on the 1,569 whose error is not at the final step. Each smolagents step has a reasoning field, a code field, and an observation field. In the row ``Reasoning, code removed'', we delete Markdown code blocks and code markers (\texttt{code} tags and ``Code:'' headers) from the reasoning field. The final-step rule is shown for reference.}
\label{tab:smolagents-local}
\begin{tabular}{@{}lrr@{}}
\toprule
Available information & All & Not final \\
\midrule
All step text & 91.9 & 84.0 \\
Reasoning only & 95.2 & 90.8 \\
Reasoning, code removed & 96.2 & 92.9 \\
Reasoning and code field & 93.9 & 88.1 \\
Reasoning and observation & 92.5 & 84.9 \\
All text except reasoning & 72.1 & 48.5 \\
\quad Code field only & 73.2 & 53.9 \\
\quad Observation only & 58.4 & 16.5 \\
Final step (no text) & 52.1 & 0.0 \\
\bottomrule
\end{tabular}\par
\end{table}

%% file: latex/table/reasoning_strata.tex
\begin{table*}[!t]
\centering\footnotesize
\setlength{\tabcolsep}{3pt}
\caption{A: number of correct predictions, written as GPT-4.1 / Claude Sonnet 5, for the 53 records whose error step states a loop and for the other 257. The one record with empty reasoning at the error step is in neither column. For each judge, the two counts in a row can therefore sum to one less than its number of correct predictions on all 311 records in Table~\ref{tab:alfagent-arms}. B: accuracy (\%) on all 311 records when the judges answer Who\&When Pro's own prompt instead of ours (Appendix~\ref{app:official-pro}). That prompt asks for the faulty agent, the error step, and the error mode. Joint is the share of records on which all three answers are correct.}
\label{tab:reasoning-strata}
\label{tab:alfagent-official}
\begin{minipage}[t]{.49\textwidth}
\centering
\textbf{A. Reasoning strata}\par\smallskip
\begin{tabular}{@{}lrr@{}}
\toprule
Input version & Loop records (53) & Other records (257) \\
\midrule
Original & 8 / 10 & 175 / 168 \\
Field-neutral & 7 / 14 & 170 / 169 \\
No reasoning & 0 / 2 & 46 / 45 \\
Template continued & 12 / 9 & 178 / 161 \\
\bottomrule
\end{tabular}\par
\end{minipage}\hfill
\begin{minipage}[t]{.49\textwidth}
\centering
\textbf{B. Benchmark's own prompt}\par\smallskip
\begin{tabular}{@{}lrrrr@{}}
\toprule
Judge & Agent & Step & Mode & Joint \\
\midrule
GPT-4.1 & 100.0 & 50.8 & 30.2 & 25.1 \\
Claude Sonnet 5 & 97.4 & 51.4 & 26.7 & 22.8 \\
\bottomrule
\end{tabular}\par
\end{minipage}

\end{table*}

%% file: latex/table/alfagent_arms.tex
\begin{table*}[!t]
\centering\footnotesize
\setlength{\tabcolsep}{3pt}
\caption{ALFAgent interventions on all 311 records. The table has three blocks. \emph{Accuracy}: exact accuracy (\%). \emph{Change from Original}: paired change in points, with 95\% task-group bootstrap intervals. Changes are computed from counts out of 311, so they can differ by 0.1 from the difference of the rounded accuracies. \emph{Boundary following}: the old boundary is the last step with reasoning under Original. The block starts from each judge's correct predictions at the old boundary under Original (183 for GPT-4.1, 178 for Claude Sonnet 5, and 175 for o3). It counts how many of them move to step 30 and how many stay at the old boundary under Template continued (rows marked Template) and under Task continued (rows marked Task). ---: not run. Paraphrase results are in Table~\ref{tab:paraphrase-results}.}
\label{tab:alfagent-arms}
\label{tab:alfagent-paired}
\label{tab:alfagent-task-related}
\begin{tabular}{@{}lrrr@{}}
\toprule
 & GPT-4.1 & Claude Sonnet 5 & o3 \\
\midrule
\multicolumn{4}{@{}l}{\textbf{Accuracy}} \\
Original & 58.8 & 57.2 & 56.3 \\
Field-neutral & 57.2 & 58.8 & 59.5 \\
No reasoning & 15.1 & 15.1 & 10.6 \\
Template continued & 61.1 & 54.7 & 60.8 \\
Task continued & 54.7 & 48.2 & --- \\
\midrule
\multicolumn{4}{@{}l}{\textbf{Change from Original}} \\
Field-neutral & $-1.6$ [$-5.2$, $+2.0$] & $+1.6$ [$-1.3$, $+4.4$] & $+3.2$ [$-1.7$, $+8.3$] \\
No reasoning & $-43.7$ [$-49.4$, $-38.1$] & $-42.1$ [$-48.1$, $-36.4$] & $-45.7$ [$-52.3$, $-39.1$] \\
Template continued & $+2.3$ [$-1.0$, $+5.4$] & $-2.6$ [$-6.5$, $+0.9$] & $+4.5$ [$+0.3$, $+8.6$] \\
Task continued & $-4.2$ [$-8.4$, $0.0$] & $-9.0$ [$-13.2$, $-5.0$] & --- \\
\midrule
\multicolumn{4}{@{}l}{\textbf{Boundary following} (of 183 / 178 / 175)} \\
Template: moved to step 30 & 0 & 0 & 1 \\
Template: stayed at old boundary & 175 & 159 & 156 \\
Task: moved to step 30 & 0 & 0 & --- \\
Task: stayed at old boundary & 155 & 139 & --- \\
\bottomrule
\end{tabular}
\end{table*}

\begin{table}[!t]
\centering\footnotesize
\setlength{\tabcolsep}{4pt}
\caption{Where wrong ALFAgent predictions land under Original: offset of the predicted step from the annotated step. Wrong predictions after the error have median offset $+8$ for both judges.}
\label{tab:alfagent-errors}
\begin{tabular}{@{}lrr@{}}
\toprule
Predicted $-$ annotated step & GPT-4.1 & Claude Sonnet 5 \\
\midrule
$\leq -6$ & 19 & 9 \\
$-5$ to $-3$ & 7 & 11 \\
$-2$ to $-1$ & 15 & 8 \\
\quad Before the error (total) & 41 & 28 \\
\addlinespace
$0$ (correct) & 183 & 178 \\
\addlinespace
$+1$ to $+2$ & 18 & 14 \\
$+3$ to $+5$ & 16 & 26 \\
$+6$ to $+10$ & 25 & 32 \\
$\geq +11$ & 28 & 33 \\
\quad After the error (total) & 87 & 105 \\
\bottomrule
\end{tabular}
\end{table}

%% file: latex/table/finetuned_judge.tex
\begin{table*}[!t]
\centering\footnotesize
\setlength{\tabcolsep}{3pt}
\caption{Judges fine-tuned on ALFAgent labels with two seeds, next to the same models before fine-tuning (Base) (Appendix~\ref{app:finetuned}). Each tuned model is trained on the Original input of four task-disjoint folds (Appendix~\ref{app:independent-step}) and tested on the held-out fold. A: exact accuracy (\%) on all 311 records, except that rows marked $^{*}$ use the 285 records of the paraphrase test. On these 285 records, Original accuracy is 99.6\% for both Seed 1 models and 50.2\% for Qwen3-8B before fine-tuning. Template before error appends the Template continued sentence to every step before the error, which keeps the reasoning boundary at the error. Boundary following: the first row counts the model's correct predictions at the old boundary (the last step with reasoning under Original). The rows below count how many of them move to step 30 and how many stay at the old boundary under Template continued and under Task continued. B: paired change from Original in points, with 95\% task-group bootstrap intervals. Changes are computed from counts, so they can differ by 0.1 from the difference of the rounded accuracies. ---: not run.}
\label{tab:finetuned}
\begin{tabular}{@{}lrrrrrr@{}}
\toprule
\multicolumn{7}{@{}l}{\textbf{A. Accuracy and boundary following}} \\
 & \multicolumn{3}{c}{Qwen2.5-1.5B} & \multicolumn{3}{c}{Qwen3-8B} \\
\cmidrule(lr){2-4}\cmidrule(l){5-7}
 & Base & Seed 0 & Seed 1 & Base & Seed 0 & Seed 1 \\
\midrule
Original & 5.5 & 100.0 & 99.7 & 51.4 & 99.7 & 99.7 \\
Field-neutral & 6.8 & 97.7 & 96.1 & 45.3 & 99.4 & 98.4 \\
No reasoning & 4.8 & 8.4 & 8.4 & 9.6 & 8.7 & 15.8 \\
Template continued & 5.1 & 70.4 & 40.5 & 50.5 & 86.5 & 98.4 \\
Template before error & --- & --- & 100.0 & 53.1 & --- & 99.7 \\
Task continued & 6.1 & 2.3 & 0.6 & 46.9 & 64.0 & 68.8 \\
Gold paraphrased$^{*}$ & --- & --- & 99.6 & 51.2 & --- & 99.6 \\
Prior paraphrased$^{*}$ & --- & --- & 100.0 & 48.1 & --- & 98.9 \\
\midrule
Boundary following: of & 17 & 310 & 309 & 159 & 309 & 309 \\
\quad Template: moved to step 30 & 0 & 0 & 0 & 0 & 0 & 0 \\
\quad Template: stayed & 11 & 218 & 125 & 145 & 268 & 305 \\
\quad Task: moved to step 30 & 1 & 26 & 0 & 1 & 7 & 7 \\
\quad Task: stayed & 8 & 7 & 1 & 131 & 198 & 213 \\
\bottomrule
\end{tabular}\par\medskip
\begin{tabular}{@{}lrrrr@{}}
\toprule
\multicolumn{5}{@{}l}{\textbf{B. Change from Original, fine-tuned models}} \\
 & \multicolumn{2}{c}{Qwen2.5-1.5B} & \multicolumn{2}{c}{Qwen3-8B} \\
\cmidrule(lr){2-3}\cmidrule(l){4-5}
 & Seed 0 & Seed 1 & Seed 0 & Seed 1 \\
\midrule
Field-neutral & $-2.3$ [$-4.0$, $-0.9$] & $-3.5$ [$-5.5$, $-1.7$] & $-0.3$ [$-1.4$, $+0.7$] & $-1.3$ [$-2.6$, $-0.3$] \\
No reasoning & $-91.6$ [$-94.8$, $-88.1$] & $-91.3$ [$-94.6$, $-87.8$] & $-91.0$ [$-94.4$, $-87.5$] & $-83.9$ [$-88.4$, $-79.0$] \\
Template continued & $-29.6$ [$-36.8$, $-22.7$] & $-59.2$ [$-68.6$, $-49.4$] & $-13.2$ [$-17.8$, $-8.8$] & $-1.3$ [$-2.6$, $-0.3$] \\
Template before error & --- & $+0.3$ [$0.0$, $+1.0$] & --- & $0.0$ [$0.0$, $0.0$] \\
Task continued & $-97.7$ [$-99.4$, $-95.8$] & $-99.0$ [$-100.0$, $-97.4$] & $-35.7$ [$-42.4$, $-29.1$] & $-30.9$ [$-36.2$, $-25.5$] \\
Gold paraphrased$^{*}$ & --- & $0.0$ [$-1.1$, $+1.0$] & --- & $0.0$ [$0.0$, $0.0$] \\
Prior paraphrased$^{*}$ & --- & $+0.4$ [$0.0$, $+1.1$] & --- & $-0.7$ [$-1.8$, $0.0$] \\
\bottomrule
\end{tabular}\par
\end{table*}

%% file: latex/appendix/aftraj.tex
\section{AFTraj: Protocol and Additional Results}
\label{app:aftraj}

\subsection{Test and models}
\label{app:aftraj-setup}

AFTraj's official test, the first-alarm test, has 163 failures. Two tasks each contribute two of these failures, so the 163 failures form 161 task groups (Appendix~\ref{app:repro}). Intervals on this test resample these task groups. AFTraj labels each trajectory as either failed, with an annotated error turn, or \emph{safe}, with no error. The test also contains 169 safe trajectories. They have no error turn to find, so they do not count toward accuracy.

A method reads a trajectory one turn (step) at a time. After each turn it sees everything up to that turn, which we call the \emph{prefix}, and decides whether to raise an alarm. At each turn, a judge either answers that the prefix is safe or names the turn that holds the error. Naming a turn counts as an alarm. The judge's prediction is the turn it names in its first alarm. A classifier gives each turn a score. Its prediction is the first turn whose score passes a threshold. A trajectory with no alarm counts as a miss. A method is right on a failure when its prediction is the annotated error turn.

We compare six \emph{methods}. Four are zero-shot judges: GPT-4.1, DeepSeek V4 Pro, Qwen3-8B, and Qwen3-32B. Two are trained classifiers: current-turn TF--IDF and Probe.

All four judges run at temperature 0 and write at most 2,048 tokens per answer. The two Qwen3 judges run with thinking mode disabled. GPT-4.1 and DeepSeek V4 Pro run with their providers' default reasoning settings.

The two classifiers differ in what they see. Current-turn TF--IDF sees only the words of the current turn, as 20,000 unigram and bigram TF--IDF features with sublinear term frequency. Probe passes the prefix through a frozen Qwen3-8B and reads the hidden state of the prefix's last token at layer 16. Both classifiers are logistic regressions trained for at most 2,000 iterations.

The classifiers learn from AFTraj trajectories whose tasks do not appear in the official test. We divide these trajectories into two sets by task group. A random 30\% of the task groups (seed 0) form a development set of 549 trajectories, which include both safe and failed ones. The 1,264 trajectories of the remaining task groups (626 safe, 421 injected, and 217 natural) train both classifiers. The development set chooses each classifier's alarm threshold. It also chooses the Qwen3-8B layer that Probe reads, from layers 8, 12, 16, 20, 24, and 28. The judges are not trained on any trajectory. Every comparison between a classifier and a judge therefore sets a trained method against a zero-shot one. Both classifiers read only the first 24 turns of a trajectory, so the two test failures whose error comes after the first 24 turns count as misses for both.

Position alone finds about a quarter of the failures, half as many as current-turn TF--IDF. As a reference we use a fixed-position predictor, which ignores the trajectory and always alarms at the same turn. The turn must not be chosen on the failures it is scored on. We therefore choose the best turn on one half of the 161 task groups, score it on the other half, and repeat with the halves swapped. Both rounds choose turn 3, which is the fourth turn because AFTraj numbers turns from 0. Always alarming at turn 3 is right on 42 of the 163 failures (25.8\%), against 82 for current-turn TF--IDF. For comparison, always alarming at turn 0 is right on none of them. Always alarming at turn 1 and always alarming at the final turn are each right on 16 of them (9.8\%).

\subsection{Residual accuracy}
\label{app:aftraj-beyond}

\textbf{GPT-4.1 and current-turn TF--IDF tie at 82 of the 163 failures, but they are right on different failures}: they share only 36. The main text measures GPT-4.1's residual accuracy against current-turn TF--IDF. GPT-4.1 alone is right on 46 failures, and current-turn TF--IDF alone is right on another 46 failures. Both are wrong on the remaining 35 failures. Current-turn TF--IDF therefore misses $46+35=81$ failures, and GPT-4.1 is right on 46 of them. GPT-4.1's residual accuracy is 46 of 81, or 56.8\% (95\% task-group bootstrap interval [45.8, 67.6]).

\input{latex/table/aftraj_overlap.tex}

Removing the failures that position alone finds leaves GPT-4.1 close to current-turn TF--IDF and Probe close to DeepSeek V4 Pro (Table~\ref{tab:aftraj-overlap}). Always guessing the fourth turn is right on 42 of the 163 failures, so part of any method's accuracy could come from position alone. Table~\ref{tab:aftraj-overlap} therefore looks only at the other 121 failures, where that guess is wrong.

First, GPT-4.1 and current-turn TF--IDF stay close. Each is right on 82 of the 163 failures. On the 121, GPT-4.1 is right on 58 (47.9\%) and current-turn TF--IDF on 54 (44.6\%).

Second, Probe and DeepSeek V4 Pro tie: Probe's overall lead of two failures (101 against 99) comes entirely from failures that the fourth-turn guess also gets right. On the other 121, each is right on 70 (57.9\%).

\subsection{Scores by failure source and domain}
\label{app:source}

\input{latex/table/source_scores.tex}
\input{latex/figure/domains_app}

Both trained classifiers are far more accurate on injected failures than on natural ones, by 45.7 points for current-turn TF--IDF and 46.9 points for Probe (Table~\ref{tab:source-scores}A, which covers the 108 injected and the 55 natural failures of the first-alarm test). Probe is the most accurate method on injected failures (77.8\%), and Qwen3-8B is the most accurate method on natural failures (58.2\%). By domain (Figure~\ref{fig:domains-app}), Probe's advantage over current-turn TF--IDF is largest in coding.

\input{latex/figure/aggregation.tex}

\textbf{The first-alarm test cannot rank its two strongest methods, Probe (62.0\%) and DeepSeek V4 Pro (60.7\%).} On injected failures, Probe is 13.9 points more accurate than DeepSeek V4 Pro [1.9, 25.5]. On natural failures, Probe is 23.6 points less accurate [10.2, 37.3]. Suppose injected failures receive weight $w$ and natural failures receive the remaining weight $1-w$. Computed from the unrounded differences, Probe's lead over DeepSeek V4 Pro in points is then
\begin{equation}
\Delta_w=13.89w-23.64(1-w).
\label{eq:leader}
\end{equation}
Probe ranks first when $w$ exceeds the crossing weight, the value of $w$ at which $\Delta_w=0$. The crossing weight is 63.0\% [38.1, 92.0] (Figure~\ref{fig:aggregation}). In 82 of the 10,000 resamples, one of the two methods is more accurate on both sources, so the crossing weights of these resamples lie outside 0--100\%. The interval [38.1, 92.0] is computed over all 10,000 crossing weights, including these 82.

Neither the released mix nor the mix of the full corpus settles the ranking, because both injected shares lie inside the crossing interval. The released test has 66.3\% injected failures. At this share, Probe is right on 101 failures and DeepSeek V4 Pro is right on 99 failures, so Probe leads by 1.2 points. The interval of this lead includes zero both when each resample pools all failures, $[-8.0, +10.6]$, and when it fixes the injected share at 66.3\%, $[-7.8, +10.2]$. With equal weights on injected and natural failures, the order reverses, and DeepSeek V4 Pro leads by 4.9 points. The full AFTraj corpus has 728 injected and 386 natural failures, so 65.4\% of its failures are injected. Its 1,162 safe trajectories do not count toward this share.

Within the agentic domain, the classifiers' injected-minus-natural accuracy gap shrinks from about 46 points to 12.5 (current-turn TF--IDF) and 8.5 (Probe), and no injected-minus-natural interval excludes zero. We look within one domain because failure source is confounded with domain: every coding failure is injected. Table~\ref{tab:aftraj-agentic}B therefore compares injected and natural failures within the agentic domain alone. It uses the same first-alarm predictions on the test's 68 agentic failures, which are 22 injected and 46 natural failures in 66 task groups. For five of the six methods, the injected-minus-natural difference has the same sign on these 68 failures as on all 163. The exception is DeepSeek V4 Pro, whose difference is $+9.3$ points on all 163 failures and $-8.9$ points on the 68 agentic failures. On the 46 natural agentic failures, a zero-shot judge beats both classifiers: Qwen3-8B is 21.7 points more accurate than Probe $[4.7, 38.6]$ and 34.8 points more accurate than current-turn TF--IDF $[14.3, 54.5]$.

\subsection{Recognizing a failing turn from its own text}
\label{app:current-turn}

Trained and tested on one failure source at a time, a classifier that reads a turn's words finds injected errors better than one that sees only the turn's structure, but it does not find natural errors better. The first-alarm test cannot show this directly, for two reasons. First, its classifiers are trained on injected and natural failures together, so their scores mix what they learn from the two sources. Second, a first-alarm prediction is wrong whenever an earlier turn passes the threshold, even if the error turn itself scores high. The experiment in this section removes both obstacles. It uses the word classifier and the structure classifier of Section~\ref{sec:population}, and each is trained and tested on one source of failure at a time. Each classifier predicts the highest-scoring turn of the trajectory rather than the first turn above a threshold. Because of this different prediction rule, the accuracies below cannot be compared with those of the first-alarm test.

\input{latex/table/current_turn.tex}

The experiment uses every agentic failure in the AFTraj corpus whose error lies within the first 24 turns. There are 468 such failures, including all 68 agentic failures of the first-alarm test. Of these, 148 are injected failures from 148 task groups, and 320 are natural failures from 253 task groups. Our copy of the corpus stores each turn as the text of the whole prefix up to that turn. In this text, each turn begins with a role marker, a line such as \texttt{[Manager]} that names the role of whoever produced the turn. We cut each prefix at its last role marker to keep only the current turn's own text. Comparing each prefix with the one before it confirms that the cut keeps no text from earlier turns.

Each classifier is trained to tell the annotated error turn of a failed trajectory from its other turns. The word classifier reads the text of a turn, with the same TF--IDF features and logistic regression as current-turn TF--IDF (Appendix~\ref{app:aftraj-setup}). The structure classifier uses the same logistic regression on six features of a turn and, apart from its role and two fixed markers, sees none of the turn's words. The first feature is the turn's role, such as \texttt{user}, \texttt{environment}, or \texttt{Manager}. The second is the turn's relative position $t/(T-1)$, where $t$ is the turn's number counted from 0 and $T$ is the number of turns the classifier reads, which is the trajectory's length capped at 24. The third is $T/24$, the capped length divided by 24. The fourth is the turn's length in characters, capped at 4,000 and divided by 4,000. The last two record whether the turn's text contains \texttt{thought:} and whether it contains \texttt{action:}. In each turn's text, these markers introduce the agent's reasoning and its action. Both classifiers use random state 0.

Each classifier is trained and tested by five-fold cross-validation within one source, so every failure is scored by a classifier trained on the other four folds. The folds keep task groups together and are formed after shuffling the groups with seed 0. When two turns get the same top score, the classifier predicts the later turn. Intervals are 95\% task-group bootstrap intervals from 10,000 resamples with seed 1.

\textbf{On injected failures, a turn's words beat its structure by 19.6 points (64.2\% against 44.6\%); on natural failures, they do not (58.1\% against 63.1\%)} (Table~\ref{tab:current-turn}). These are the rates at which the word classifier and the structure classifier find the annotated error turn. As a reference, always predicting one fixed turn finds it in 40.5\% of injected failures and 34.4\% of natural failures.

%% file: latex/table/aftraj_overlap.tex
\begin{table}[!t]
\centering\scriptsize
\setlength{\tabcolsep}{4pt}
\caption{Accuracy on the 121 failures that the fixed-position predictor gets wrong. The predictor always answers the fourth turn and is right on 42 of the 163 failures. \emph{Correct}: failures a method gets right, out of 163. \emph{Predictor also correct}: how many of those the predictor also gets right. \emph{Among the 121}: Correct minus Predictor also correct, i.e., the failures the method gets right and the predictor gets wrong. \emph{Acc.\ on the 121}: that count divided by 121 (residual accuracy against the fixed-position predictor).}
\label{tab:aftraj-overlap}
\begin{tabular}{@{}lrrrr@{}}
\toprule
 & Correct & \shortstack[r]{Predictor\\also correct} & \shortstack[r]{Among\\the 121} & \shortstack[r]{Acc.\ on the\\121 (\%)} \\
\midrule
Qwen3-8B & 73 & 17 & 56 & 46.3 \\
Qwen3-32B & 57 & 16 & 41 & 33.9 \\
GPT-4.1 & 82 & 24 & 58 & 47.9 \\
DeepSeek V4 Pro & 99 & 29 & 70 & 57.9 \\
Current-turn TF--IDF & 82 & 28 & 54 & 44.6 \\
Probe & 101 & 31 & 70 & 57.9 \\
\bottomrule
\end{tabular}
\end{table}

%% file: latex/table/source_scores.tex
\begin{table*}[!t]
\centering\footnotesize
\setlength{\tabcolsep}{4pt}
\caption{AFTraj first-alarm accuracy (\%) on injected and on natural failures. Panel A covers all 163 failures of the first-alarm test. Panel B covers the 68 agentic failures among them and uses the same predictions. The last column is injected minus natural accuracy, in points. Rows under \emph{Between methods} give the difference between two methods' accuracies on the same failures, in points. Brackets are 95\% task-group bootstrap intervals. DeepSeek is DeepSeek V4 Pro, and TF--IDF is current-turn TF--IDF. Bold marks the best method in a column. In the Released column, the interval for Probe $-$ DeepSeek pools all failures in each resample, so the share of injected failures varies from resample to resample. If each resample instead weights its injected and natural differences by the released shares, 108/163 and 55/163, the interval is $[-7.8, +10.2]$.}
\label{tab:source-scores}
\label{tab:aftraj-agentic}
\textbf{A. First-alarm test: 108 injected, 55 natural}\par\smallskip
\begin{tabular}{@{}lrrrr@{}}
\toprule
Method & Released & Injected & Natural & Injected $-$ natural \\
\midrule
Qwen3-8B & 44.8 & 38.0 & \textbf{58.2} & $-20.2$ [$-36.7$, $-5.0$] \\
Qwen3-32B & 35.0 & 26.9 & 50.9 & $-24.1$ [$-39.8$, $-8.1$] \\
GPT-4.1 & 50.3 & 48.1 & 54.5 & $-6.4$ [$-21.9$, $+10.5$] \\
DeepSeek V4 Pro & 60.7 & 63.9 & 54.5 & $+9.3$ [$-6.3$, $+25.6$] \\
Current-turn TF--IDF & 50.3 & 65.7 & 20.0 & $+45.7$ [$+31.4$, $+58.6$] \\
Probe & \textbf{62.0} & \textbf{77.8} & 30.9 & $+46.9$ [$+31.2$, $+61.5$] \\
\midrule
\multicolumn{5}{@{}l}{\emph{Between methods (paired differences)}} \\
Probe $-$ DeepSeek & $+1.2$ [$-8.0$, $+10.6$] & $+13.9$ [$+1.9$, $+25.5$] & $-23.6$ [$-37.3$, $-10.2$] & \\
TF--IDF $-$ GPT-4.1 & 0.0 & $+17.6$ [$+4.0$, $+31.5$] & $-34.5$ [$-52.0$, $-15.4$] & \\
\bottomrule
\end{tabular}\par
\medskip
\textbf{B. Agentic domain: 22 injected, 46 natural}\par\smallskip
\begin{tabular}{@{}lrrr@{}}
\toprule
Method & Injected & Natural & Injected $-$ natural \\
\midrule
Qwen3-8B & 40.9 & \textbf{58.7} & $-17.8$ [$-42.9$, $+8.2$] \\
Qwen3-32B & \textbf{50.0} & 54.3 & $-4.3$ [$-30.8$, $+21.5$] \\
GPT-4.1 & 40.9 & 54.3 & $-13.4$ [$-39.0$, $+12.5$] \\
DeepSeek V4 Pro & 45.5 & 54.3 & $-8.9$ [$-34.8$, $+17.5$] \\
Current-turn TF--IDF & 36.4 & 23.9 & $+12.5$ [$-11.6$, $+36.8$] \\
Probe & 45.5 & 37.0 & $+8.5$ [$-17.4$, $+34.3$] \\
\midrule
\multicolumn{4}{@{}l}{\emph{Between methods (paired differences, natural failures only)}} \\
Qwen3-8B $-$ Probe & & $+21.7$ [$+4.7$, $+38.6$] & \\
Qwen3-8B $-$ TF--IDF & & $+34.8$ [$+14.3$, $+54.5$] & \\
\bottomrule
\end{tabular}
\end{table*}

%% file: latex/figure/domains_app.tex
\begin{figure}[!htbp]
\centering
\includegraphics[width=\columnwidth]{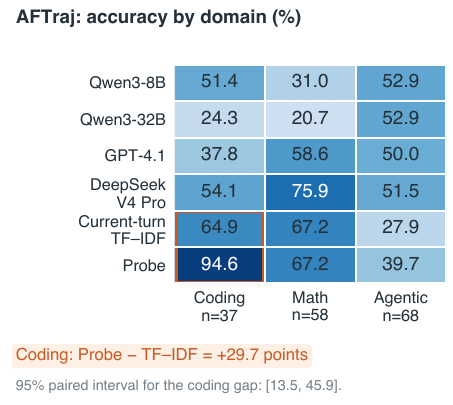}
\caption{AFTraj accuracy by domain on the first-alarm test. Probe's advantage over current-turn TF--IDF is largest in coding, and the orange outline marks this coding comparison. On the 68 agentic failures, Qwen3-8B and Qwen3-32B each score 52.9\%, DeepSeek V4 Pro scores 51.5\%, and GPT-4.1 scores 50.0\%.}
\label{fig:domains-app}
\end{figure}

%% file: latex/figure/aggregation.tex
\begin{figure}[t]
\centering
\includegraphics[width=\columnwidth]{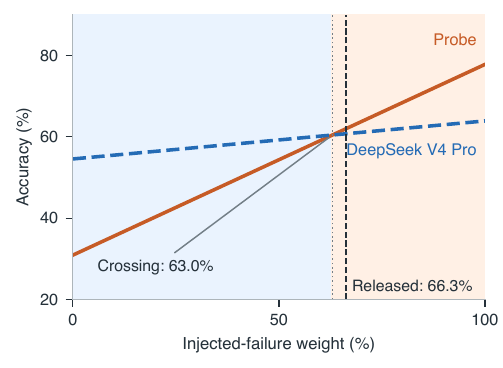}
\caption{Accuracy of Probe and DeepSeek V4 Pro when the same first-alarm predictions are reweighted, with weight $w$ on injected failures and $1-w$ on natural failures. Probe leads DeepSeek V4 Pro once injected failures carry more than 63.0\% of the weight. The vertical dashed line marks the released mixture (66.3\% injected), and the vertical dotted line marks the crossing weight (63.0\%).}
\label{fig:aggregation}
\end{figure}

%% file: latex/table/current_turn.tex
\begin{table}[!t]
\centering\footnotesize
\setlength{\tabcolsep}{4pt}
\caption{How often a predictor picks the annotated error turn, on 468 agentic failures from the AFTraj corpus (\%, with 95\% task-group bootstrap intervals). Column heads give the number of failures. Each classifier picks its highest-scoring turn and is trained and tested on injected or on natural failures separately. The fixed-position predictor always predicts one turn, chosen by cross-validation on folds that mix injected and natural failures. These accuracies are not comparable to those of the first-alarm test.}
\label{tab:current-turn}
\begin{tabular}{@{}lrr@{}}
\toprule
Predictor & Injected (148) & Natural (320) \\
\midrule
Word classifier & 64.2 [56.8,71.6] & 58.1 [51.8,64.6] \\
Structure classifier & 44.6 [36.5,52.7] & 63.1 [55.9,70.1] \\
Fixed position & 40.5 [33.1,48.6] & 34.4 [28.3,41.1] \\
\bottomrule
\end{tabular}
\end{table}

%% file: latex/appendix/aegis.tex
\section{Aegis: Predictors, Judge Run, and Data Release}
\label{app:aegis}
\label{app:aegis-paired}

\subsection{Restricted-input predictors}

\input{latex/table/aegis_full.tex}

The restricted-input predictors see only the sequence of speaker names in a conversation, one per turn, without any message text. We consider three. \emph{All speakers} labels every speaker faulty. \emph{Final speaker} labels only the speaker of the last turn faulty. \emph{Most frequent} labels only the speaker with the most turns faulty, and it breaks ties alphabetically.

As described in Appendix~\ref{app:repro}, we choose the predictor on one half of the 441 query groups (each group holds the conversations built from the same query), score it on the other half, and then swap the two halves. Both rounds pick the all-speakers predictor.

\textbf{All speakers wins on micro-F1 through recall alone, yet it is the worst of the three on exact sets} (Table~\ref{tab:aegis-full}). Its micro-F1 is 49.2\%, against 21.4\% for final speaker and 25.3\% for most frequent, because its recall is 70.2\% and theirs is below 20\%. All speakers names everyone, yet its recall is not 100\% because some annotated names match no listed speaker (Appendix~\ref{app:aegis-names}). Its exact-set accuracy is only 1.5\%. The benchmark's macro-F1 even ranks all speakers (18.1\%) above GPT-4.1 (15.3\%).

\subsection{Judge run and name matching}
\label{app:aegis-names}

GPT-4.1 receives the benchmark's official chain-of-thought prompt with the query and the conversation history. It does not see the labels or the dataset fields that record how each error was injected. Unlike our other GPT-4.1 runs, which use temperature 0, this run used the API default temperature. Two of the 600 outputs are invalid and count as misses.

Speaker names in the annotations do not always match the speaker list. 417 annotated name occurrences, in 147 conversations, match no listed speaker. Lowercasing and trimming whitespace fixes 275 of the 417 and leaves 142 unmatched.

We therefore report scores under two versions of the speaker names. Table~\ref{tab:aegis-full}, which compares the three predictors, uses the names as released. Table~\ref{tab:aegis-paired-app} in the main text, which compares GPT-4.1 with all speakers, gives both versions: names as released, and names lowercased and trimmed in both predictions and labels.

\subsection{Which release we audit}
\label{app:aegis-versions}

Our GPT-4.1 numbers cannot be compared with those of the original Aegis paper, because the public data has changed across revisions and that paper does not name the revision behind its judge scores. All Aegis results here use one set of 600 conversations, from Hugging Face revision \texttt{65bd10d8}. The current Hugging Face release gives every conversation a new ID, and 107 of our 600 conversations no longer appear in it. The supplementary material lists the data identifiers and explains how we matched conversations across revisions.

%% file: latex/table/aegis_full.tex
\begin{table}[!t]
\centering\footnotesize
\setlength{\tabcolsep}{4pt}
\caption{Scores (\%) of the three restricted-input predictors and of GPT-4.1 on the 600 Aegis conversations, with speaker names as released (not lowercased or trimmed). Each predictor sees only the sequence of speaker names, one per turn. Micro-F1, used to choose the predictor, pools true positives, false positives, and false negatives over all conversations. Macro-F1, from the benchmark's released scorer, averages per-name F1 over every distinct name in the annotations or the predictions.}
\label{tab:aegis-full}
\begin{tabular}{@{}lrrrr@{}}
\toprule
 & \shortstack[r]{All\\speakers} & \shortstack[r]{Final\\speaker} & \shortstack[r]{Most\\frequent} & GPT-4.1 \\
\midrule
Precision & 37.8 & 35.7 & 42.2 & 56.0 \\
Recall & 70.2 & 15.3 & 18.1 & 45.1 \\
Micro-F1 & 49.2 & 21.4 & 25.3 & 50.0 \\
Macro-F1 & 18.1 & 6.2 & 8.7 & 15.3 \\
Exact set & 1.5 & 6.3 & 6.2 & 20.7 \\
\bottomrule
\end{tabular}
\end{table}

%% file: latex/appendix/video.tex
\section{Who\&When Pro Video Position Audit}
\label{app:video}

\textbf{On Who\&When Pro's video subset, a predictor that always answers the final step beats GPT-4.1 by 3.5 points, averaged over the two systems.} The subset comes from two agent systems: DeepVideoDiscovery (DVD), which searches indexed clips, and EVA, which plans what to inspect and picks frames iteratively. We run two of the four tests on it, input restriction and residual accuracy (Section~\ref{sec:formal}). We skip cue following because our cue-following interventions edit reasoning text, and video trajectories have none.

\paragraph{Step numbers.}
Step positions, and hence the final-step predictor, follow the benchmark's released renderers, the scripts that turn each raw trajectory file into numbered steps, rather than the raw files. A step is one agent interaction. For DVD, the renderer drops two leading entries from the raw trajectory. For EVA, the renderer counts only assistant actions as steps and attaches each tool output to the preceding action as its observation.

\paragraph{Judge and predictor.}
GPT-4.1 runs in the released harness at temperature 0 with a 1,024-token output limit. The restricted-input predictor sees only step positions, with no text or frames. It always answers the same position. Five-fold cross-validation over task groups, run separately for each system, chooses this position from three candidates: the final step, the penultimate step, and the one before that. Every fold chooses the final step, so we call it the final-step predictor.

\input{latex/table/video_offset.tex}

\paragraph{Input restriction.}
Step position alone beats GPT-4.1 on video. The final-step predictor is right on 123 of 168 DVD trajectories (73.2\%) and 400 of 748 EVA trajectories (53.5\%), an average of 63.3\% over the two systems. GPT-4.1 is right on 111 DVD trajectories (66.1\%) and 401 EVA trajectories (53.6\%), an average of 59.8\%, which is 3.5 points lower.

\paragraph{Residual accuracy.}
On DVD, GPT-4.1's residual accuracy of 60.0\%, read alone, overrates GPT-4.1. The final-step predictor misses the 45 errors that lie before the final step (24 penultimate, 21 earlier; Table~\ref{tab:video-offset}), and GPT-4.1 finds 27 of them. GPT-4.1 also misses 39 of the 123 final-step errors that the predictor gets right, so GPT-4.1's DVD accuracy is 7.1 points lower than the predictor's.

On EVA, GPT-4.1 behaves almost like the predictor. Every error is at the final step (400) or the penultimate step (348). GPT-4.1 answers the final step in 744 of the 748 trajectories, so it finds only 2 of the 348 penultimate errors (0.6\%) and misses 1 of the 400 final-step errors.

%% file: latex/table/video_offset.tex
\begin{table}[ht]
\centering\footnotesize
\caption{Number of trajectories whose annotated error (row Annotation) or GPT-4.1 prediction (row GPT-4.1) lies at the final step, the penultimate step, or an earlier step. Each row sums to the system total (168 DVD, 748 EVA). The table shows positions only and does not mark whether a prediction is right.}
\label{tab:video-offset}
\setlength{\tabcolsep}{4pt}
\begin{tabular}{@{}llrrr@{}}
\toprule
System & \shortstack{Annotation /\\prediction} & Final & Penultimate & Earlier \\
\midrule
DVD & Annotation & 123 & 24 & 21 \\
 & GPT-4.1 & 84 & 30 & 54 \\
EVA & Annotation & 400 & 348 & 0 \\
 & GPT-4.1 & 744 & 2 & 2 \\
\bottomrule
\end{tabular}\par
\end{table}